\documentclass[runningheads]{llncs}
\usepackage{graphicx}
\usepackage{amsmath,amssymb} 
\usepackage{color}
\usepackage{algorithm} 
\usepackage{algorithmic} 
\usepackage{subcaption}
\begin{document}
\title{Accurate Scene Text Detection through Border Semantics Awareness and Bootstrapping} 

\titlerunning{Scene Text Detection through Border and Bootstrapping}
%
\author{Chuhui Xue\orcidID{0000-0002-3562-3094} \and
Shijian Lu\orcidID{0000-0002-6766-2506} \and
Fangneng Zhan\orcidID{0000-0003-1502-6847}}
%
\authorrunning{Chuhui Xue, Shijian Lu, Fangneng Zhan}

\institute{School of Computer Science and Engineering,
	Nanyang Technological University
    \email{xuec0003@e.ntu.edu.sg},
    \email{ \{shijian.lu,fnzhan\}@ntu.edu.sg}
}

\maketitle              
\begin{abstract}
This paper presents a scene text detection technique that exploits bootstrapping and text border semantics for accurate localization of texts in scenes. A novel bootstrapping technique is designed which samples multiple `subsections' of a word or text line and accordingly relieves the constraint of limited training data effectively. At the same time, the repeated sampling of text `subsections' improves the consistency of the predicted text feature maps which is critical in predicting a single complete instead of multiple broken boxes for long words or text lines. In addition, a semantics-aware text border detection technique is designed which produces four types of text border segments for each scene text. With semantics-aware text borders, scene texts can be localized more accurately by regressing text pixels around the ends of words or text lines instead of all text pixels which often leads to inaccurate localization while dealing with long words or text lines. Extensive experiments demonstrate the effectiveness of the proposed techniques, and superior performance is obtained over several public datasets, e. g. 80.1 f-score for the MSRA-TD500, 67.1 f-score for the ICDAR2017-RCTW, etc.
\end{abstract}

\section{Introduction} \label{sec:intro}
Scene text detection and recognition has attracted increasing interests in recent years in both computer vision and deep learning research communities due to its wide range of applications in multilingual translation, autonomous driving, etc. As a prerequisite of scene text recognition, detecting text in scenes plays an essential role in the whole chain of scene text understanding processes. Though studied for years, accurate and robust detection of texts in scenes is still a very open research challenge as witnessed by increasing benchmarking competitions in recent years such as ICDAR2015-Incidental \cite{karatzas2015icdar}, ICDAR2017-MLT \cite{nayef2017icdar2017}, etc.

With the fast development of convolutional neural networks (CNN) in representation learning and object detection, two CNN-based scene text detection approaches have been investigated in recent years which treat words or text lines as generic objects and adapt generic object detection techniques for the scene text detection task. One approach is indirect regression based \cite{Liu_2017_CVPR,liao2017textboxes,He2017Single,jiang2017r2cnn} which employs object detectors such as Faster-RCNN \cite{ren2015faster} and SSD \cite{liu2016ssd} that first generate proposals or default boxes and then regress to accurate object boxes. These techniques achieve state-of-the-art performance but require multiple proposals of different lengths, angles and shapes. Another approach is direct regression based \cite{He2017Deep,Zhou_2017_CVPR} which adapts DenseBox \cite{huang2015densebox} for the scene text detection task. This approach does not require proposals and is capable of detecting words and text lines of different orientations and lengths, but it often suffers from low localization accuracy while dealing with long words or text lines.

\begin{figure}[t] 
\centering
\includegraphics[width=0.99\textwidth]{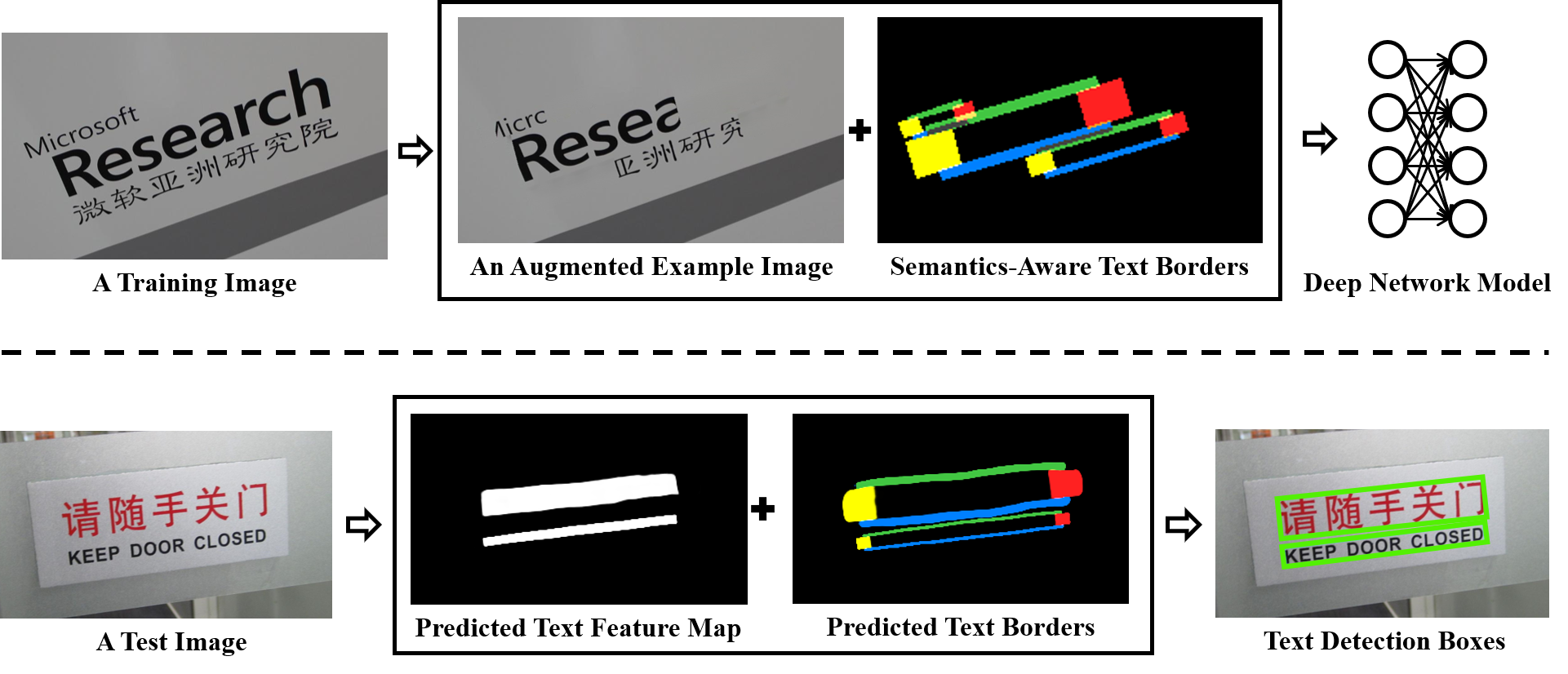}
\caption{Overview of proposed scene text detection technique: For each training image, a set of augmented images and semantics-aware text borders are extracted and fed to a multi-channel fully convolutional network to train a scene text detector (as shown above the dotted line). Given a test image, the scene text detector predicts a text feature map and four text borders (highlighted in four colors) for accurate scene text detection (as shown below the dotted line).}
\label{fig:overview}
\end{figure}

Both direct and indirect regression based approaches are thus facing three common constrains while adapted for the scene text detection task. The first is broken detections where a text line is detected as multiple broken text segments. The reason is that text lines often suffer from more variation as compared with characters or words due to their larger spatial coverage in scenes, e. g. different words within a text line may have different colors, fonts, environmental lighting, etc. The second is inaccurate localization, where the regression fails to produce an accurate text box either by missing certain parts of texts or including certain neighboring background. The inaccurate localization is largely due to the long shape of text lines where the regressing text pixels around the text line center are very far from text line ends where text bounding box vertices are located. The third is limited training data. A large amount of annotations are required to capture the rich variation within scene texts, but existing datasets often have limited training images, e.g. 300 training images in MSRA-TD500 \cite{yao2012detecting}, 229 training images in ICDAR2013 \cite{karatzas2013icdar}, etc.

We design two novel techniques to tackle the three constraints of state-of-the-art scene text detection techniques. First, we design a novel bootstrapping based scene text sampling technique that repeatedly extracts text segments of different lengths from annotated texts as illustrated in Fig. \ref{fig:overview}. The bootstrapping based sampling helps from two aspects. First, it augments the training data and relieves the data annotation constraint by leveraging existing scene text annotations. Second, the repeated sampling of text segments of various lengths helps to decouple different types of image degradation and reduce the complexity of training data effectively, e. g. scene texts with different lighting within the same text line could be sampled by different text line segments with less variation as illustrated in Fig. \ref{fig:augment}. The proposed bootstrapping based scene text sampling technique thus helps to improve the consistency of the produced text feature map and performance of regression which are critical for detecting a complete instead of multiple broken boxes for a long word or text line. The idea of repeated sampling has been exploited in training generic object detectors by cropping multiple samples around annotated objects of interest.

Second, we design a novel semantics-aware text border detection technique for accurate localization of texts in scenes. In particular, four text border segments are defined by a pair of long-side borders and a pair of short-side borders which can be extracted based on the text annotation boxes automatically as illustrated in Figs. \ref{fig:overview} and \ref{fig:labeling}. By labeling the four text border segments as four types of objects, the trained scene text detector is capable of detecting the four types of text border segments separately as illustrated in Fig. \ref{fig:overview} (four colors are for illustration only). The differentiation of the four text border segments helps to improve the text localization accuracy from two aspects. First, the text bounding box can be regressed more accurately by using text pixels lying around the two ends of words or text lines (which can be identified by using the short-side text border segments) that are much closer to the text bounding box vertices as compared with text pixels lying around the middle of text lines. Second, the long-side text border segments can be exploited to separate neighboring text lines especially when they are close to each other.

\section{Related Work} \label{sec:relate}
\subsubsection{Scene Text Detection} 
Quite a number of scene text detection techniques have been reported in the literature \cite{zhu2016scene,ye2015text} and they can be broadly classified into three categories depending on whether they detect characters, words, or text lines directly. The first category takes a bottom-up approach which first detects characters \cite{tian2015text,cho2016canny,yin2015multi} or text components \cite{tian2016detecting,Shi_2017_CVPR} and then groups them into words or text lines. The earlier works detect characters using various hand-crafted features such as stroke width transform (SWT) \cite{yao2012detecting,epshtein2010detecting}, maximally stable extremal regions (MSERs) \cite{cho2016canny,neumann2012real,huang2014robust,kang2014orientation}, boundary \cite{lu2015scene}, FAST keypoints \cite{busta1fastext}, histogram of oriented gradients (HoG) \cite{tian2015text}, stroke symmetry \cite{zhang1symmetry}, etc. With the fast development of deep neural networks, CNNs have been widely used to detect characters in scenes, either by adapting generic object detection methods \cite{tian2016detecting,Shi_2017_CVPR} or taking a semantic image segmentation approach \cite{zhang2016multi,yao2016scene,he2016text}. Additionally, different techniques have been developed to connect the detected characters into words or text lines by using TextFlow \cite{tian2015text}, long short-term memory (LSTM) \cite{zhang2016multi}, etc \cite{jaderberg2014deep,yao2016scene,liu2018mcn}.

The second category treats words as one specific type of objects and detects them directly by adapting various generic object detection techniques. The methods under this category can be further classified into two classes. The first class leverages Faster-RCNN \cite{ren2015faster}, YOLO \cite{redmon2016you} and SSD \cite{liu2016ssd} and designs text-specific proposals or default boxes for scene text detection \cite{Liu_2017_CVPR,liao2017textboxes,He2017Single,jiang2017r2cnn,Gupta_2016_CVPR,Tian_2017_ICCV}. The second class takes a direct regression approach \cite{He2017Deep,Zhou_2017_CVPR} which first detects region of interest (ROI) and then regresses text boxes around the ROI at pixel level.

The third category detects text lines directly by exploiting the full convolution network (FCN) \cite{long2015fully} that has been successfully applied for semantic image segmentation. For example, He \textit{et al.} \cite{he2016accurate} proposed a coarse-to-fine FCN that detects scene texts by extracting text regions and text central lines. In \cite{Wu_2017_ICCV,polzounov2017wordfence}, FCN is exploited to learn text border maps, where text lines are detected by finding connected components with text labels.

Our proposed technique takes the direct regression approach as in \cite{He2017Deep,Zhou_2017_CVPR} that regresses word and text line boxes directly from text pixels. On the other hand, we detect multiple text border segments with specific semantics (instead of a whole text border as in \cite{Wu_2017_ICCV,polzounov2017wordfence}) that help to improve the scene text localization accuracy greatly, more details to be described in Sec. \ref{sec:border}.
\subsubsection{Data Augmentation} 
Data augmentation has been widely adopted in deep network training as a type of regularization for avoiding over-fitting. For various computer vision tasks such as image classification and object detection, it is widely implemented through translating, rotating, cropping and flipping of images or annotated objects of interest for the purpose of creating a larger amount of training data \cite{krizhevsky2012imagenet,simonyan2014very,he2016deep}. Some more sophisticated augmentation schemes have been proposed in recent years, e. g. using masks to hide certain parts of objects to simulate various occlusion instances \cite{zhong2017random}. Data augmentation has become one routine operation in deep learning due to its effectiveness in training more accurate and more robust deep network models.

Our bootstrapping based scene text sampling falls under the umbrella of data augmentation. It is similar to image cropping but involves innovative designs by catering to text-specific shapes and structures. By decoupling image variations in long words or text lines, it helps to produce more consistent scene text features which is critical in predicting a single complete instead of multiple broken boxes for a word or text line, more details to be described in Sec. \ref{sec:aug}.

\section{Methodology} \label{sec:method}
We proposed a novel scene text detection technique that exploits bootstrapping for data augmentation and semantics-aware text border segments for accurate scene text localization. For each training image, the proposed technique extracts a set of bootstrapped training samples and two pairs of text border segments as illustrated in Fig. \ref{fig:overview}, and feeds them (together with the original scene text annotations) to a multi-channel fully convolutional network to train a scene text detection model. The bootstrapping based sampling improves the consistency of the produced text feature map which greatly helps to predict a single complete instead of multiple broken boxes for long words or text lines. The semantics of the detected text border segments greatly help to regress more accurate localization boxes for words or text lines in scenes as illustrated in Fig. \ref{fig:overview}.

\subsection{Bootstrapping based Image Augmentation} \label{sec:aug}
\begin{figure}[t]
\centering
\includegraphics[width=0.99\textwidth]{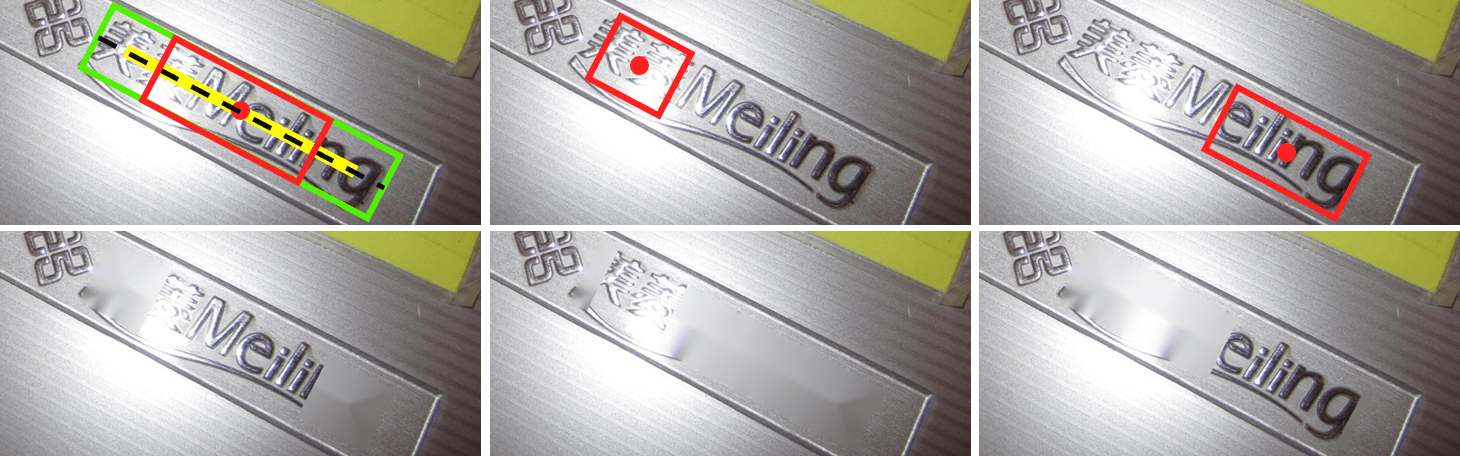}
\caption{Illustration of the bootstrapping based scene text sampling: Given an image with a text line as annotated by the green box, three example text line segments are extracted as highlighted by red boxes where the centers of the sampling windows are taken randomly along the center line of the text line (the shrunk part in yellow color). The rest text regions outside of the sampling windows are filled by inpainting.}
\label{fig:augment}
\end{figure}

We design a bootstrapping based image augmentation technique that repeatedly samples text line segments for each text annotation box (TAB) as labeled by the green box in the top-left image in Fig. \ref{fig:augment}. With $L$ denoting the TAB length, the center line of the TAB (as highlighted by the dashed line) is first shrunk by $0.1*L$ from both TAB ends which gives the yellow line segment as shown in Fig. \ref{fig:augment}. Multiple points are then taken randomly along the shrunk center line for text segment sampling. The length of each sampled text segment varies from $0.2*L$ to twice the distance between the sampling point to the closer TAB end. In addition, the rest of the TAB outside the sampled text segment is filled by inpainting [42] as illustrated in Fig. \ref{fig:augment}. With the sampling process as described above, the number of the augmented images can be controlled by the number of text segments that are sampled from each text box.

\begin{figure}[t]
\centering
\begin{subfigure}[t]{0.19\textwidth}
\includegraphics[width=\textwidth]{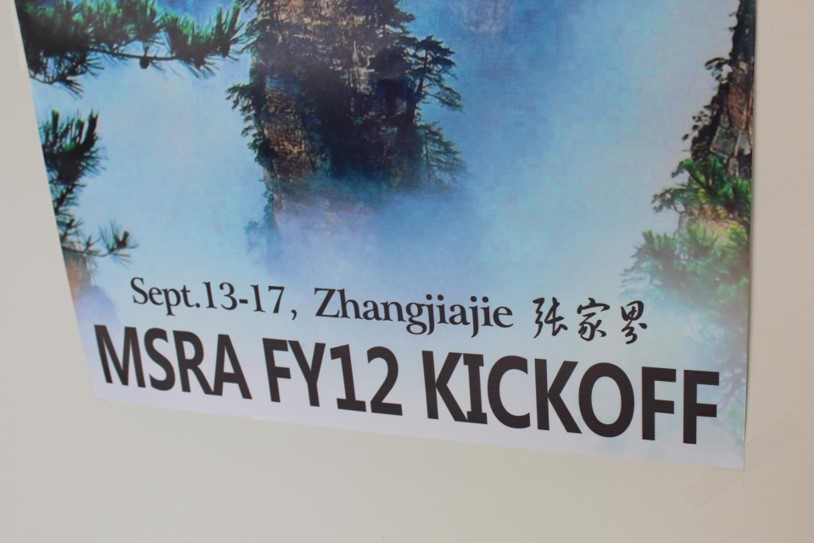}
\end{subfigure}
\begin{subfigure}[t]{0.19\textwidth}
\includegraphics[width=\textwidth]{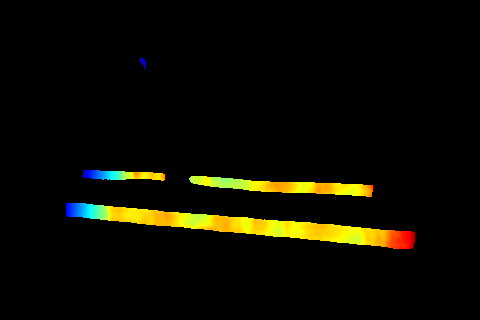}
\end{subfigure}
\begin{subfigure}[t]{0.19\textwidth}
\includegraphics[width=\textwidth]{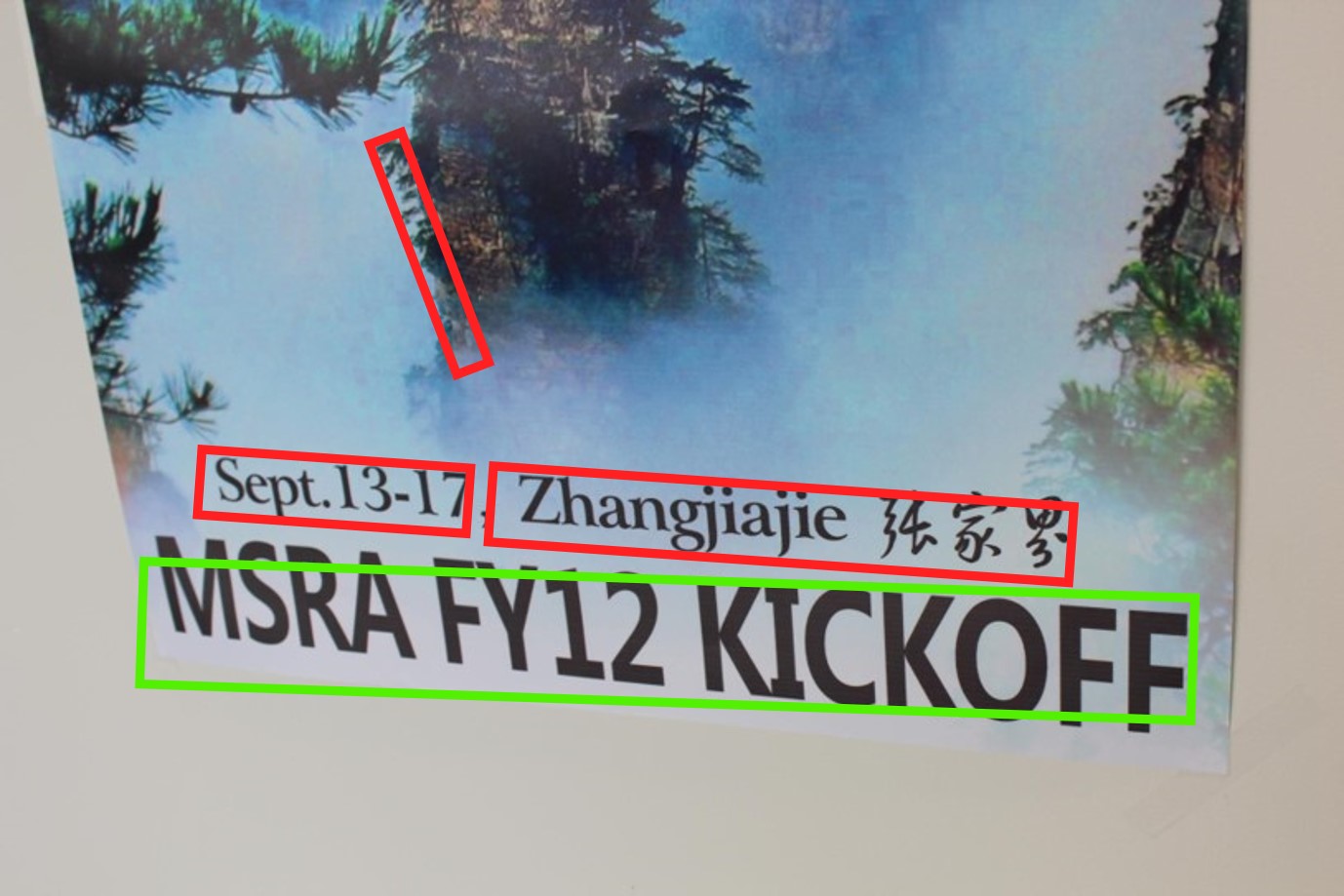}
\end{subfigure}
\begin{subfigure}[t]{0.19\textwidth}
\includegraphics[width=\textwidth]{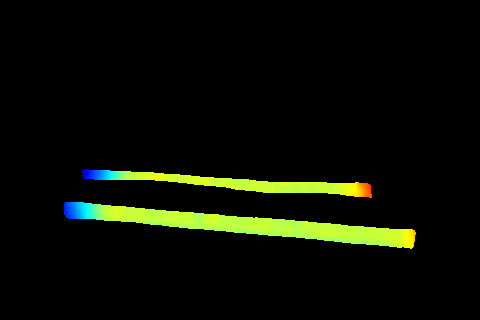}
\end{subfigure}
\begin{subfigure}[t]{0.19\textwidth}
\includegraphics[width=\textwidth]{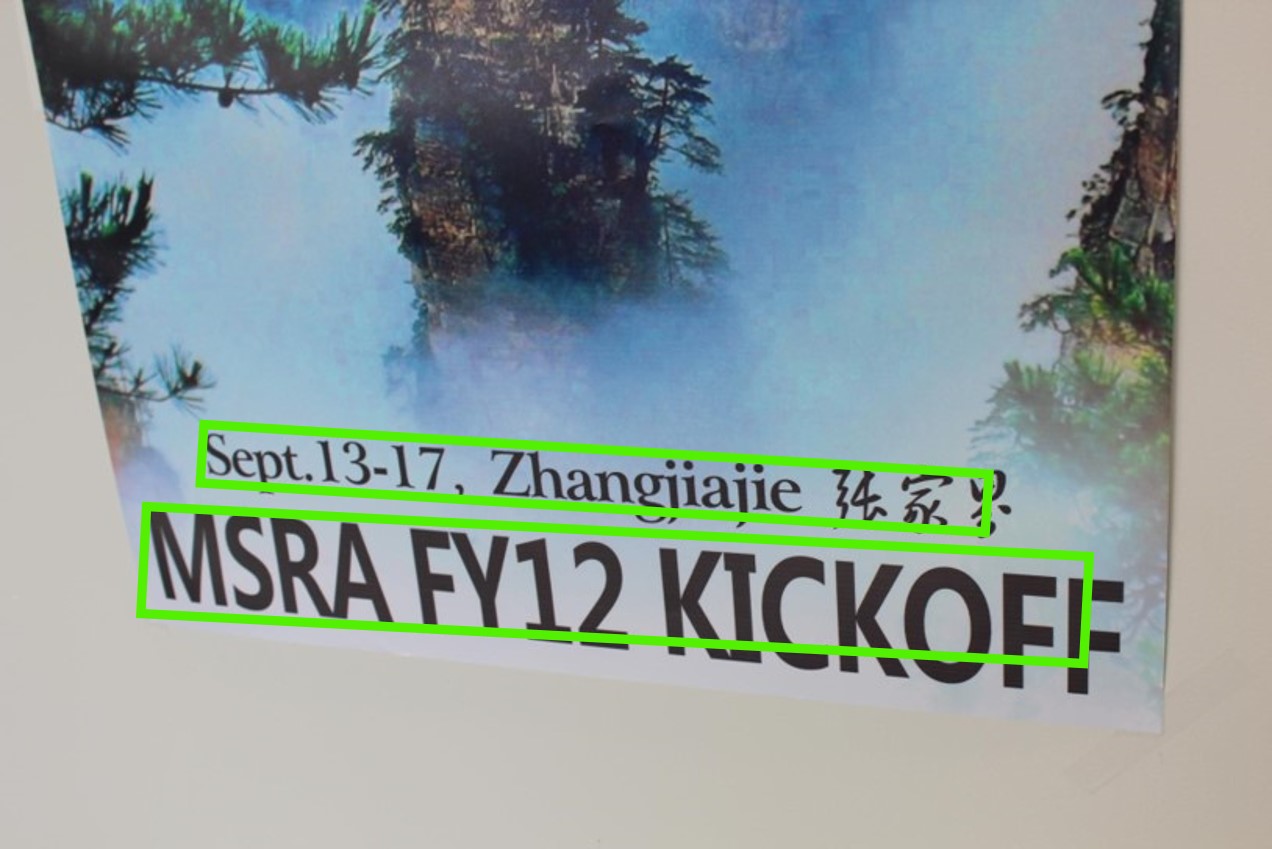}
\end{subfigure}
\begin{subfigure}[t]{0.19\textwidth}
\includegraphics[width=\textwidth]{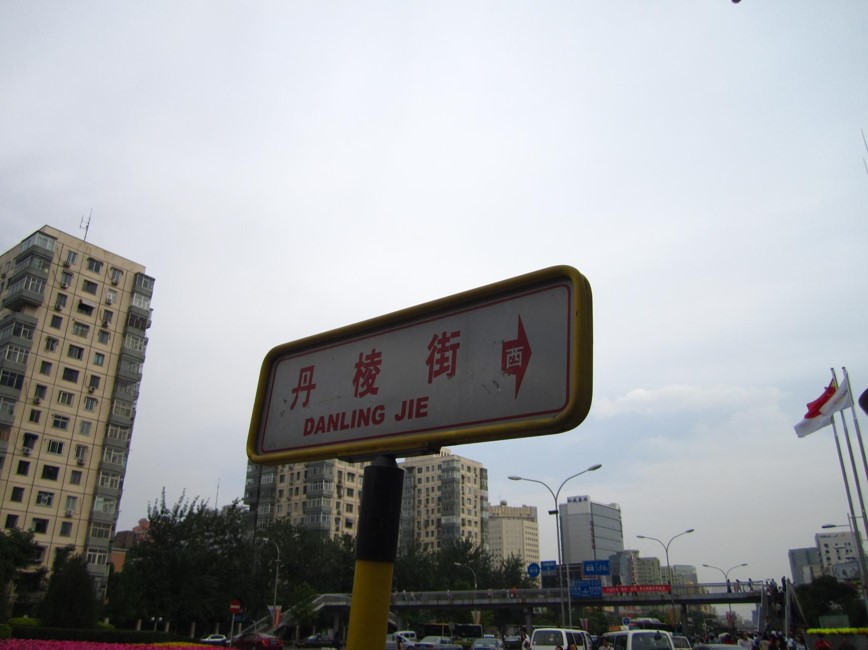}
\end{subfigure}
\begin{subfigure}[t]{0.19\textwidth}
\includegraphics[width=\textwidth]{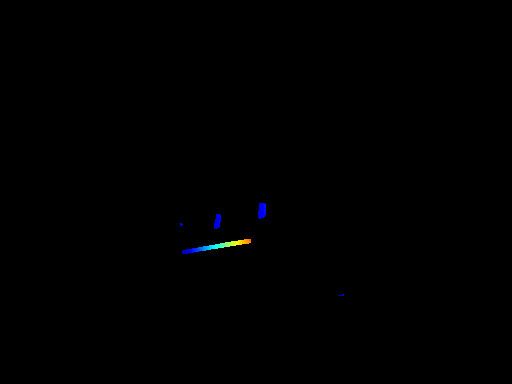}
\end{subfigure}
\begin{subfigure}[t]{0.19\textwidth}
\includegraphics[width=\textwidth]{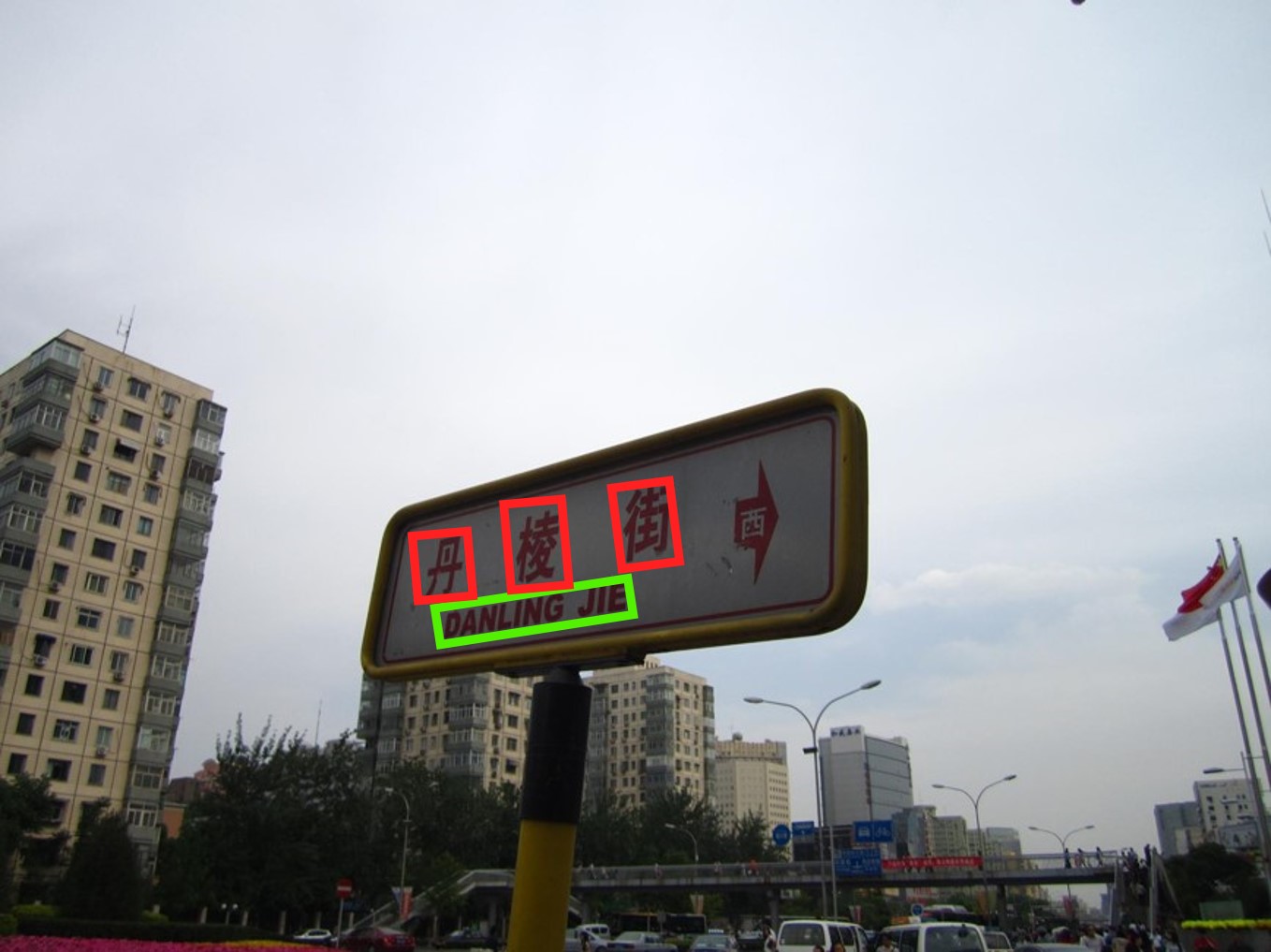}
\end{subfigure}
\begin{subfigure}[t]{0.19\textwidth}
\includegraphics[width=\textwidth]{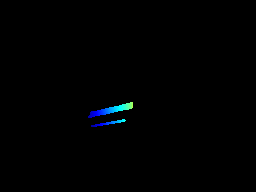}
\end{subfigure}
\begin{subfigure}[t]{0.19\textwidth}
\includegraphics[width=\textwidth]{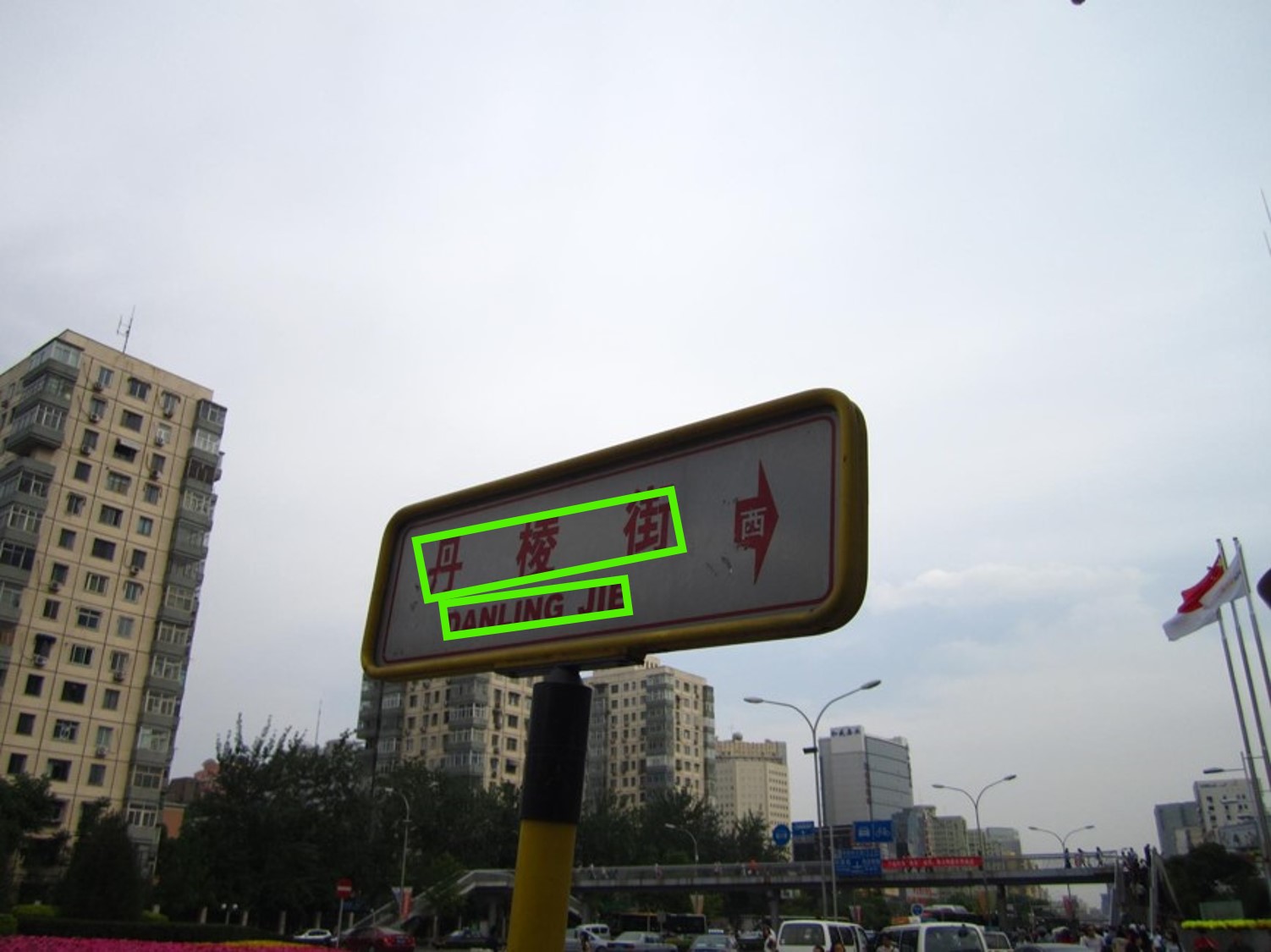}
\end{subfigure}
\begin{subfigure}[t]{0.19\textwidth}
\includegraphics[width=\textwidth]{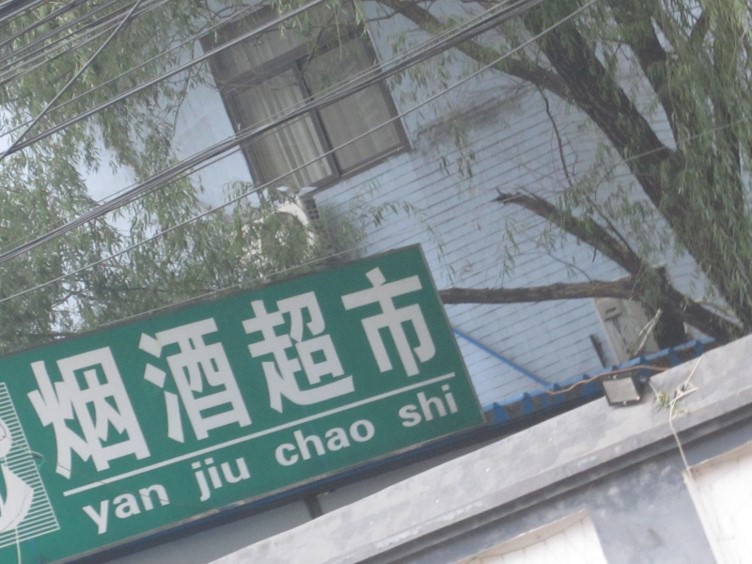}
\caption{}
\label{fig:aug_show_orig}
\end{subfigure}
\begin{subfigure}[t]{0.19\textwidth}
\includegraphics[width=\textwidth]{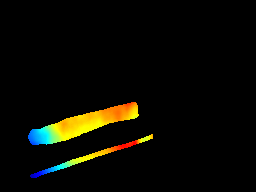}
\caption{}
\label{fig:aug_show_geo}
\end{subfigure}
\begin{subfigure}[t]{0.19\textwidth}
\includegraphics[width=\textwidth]{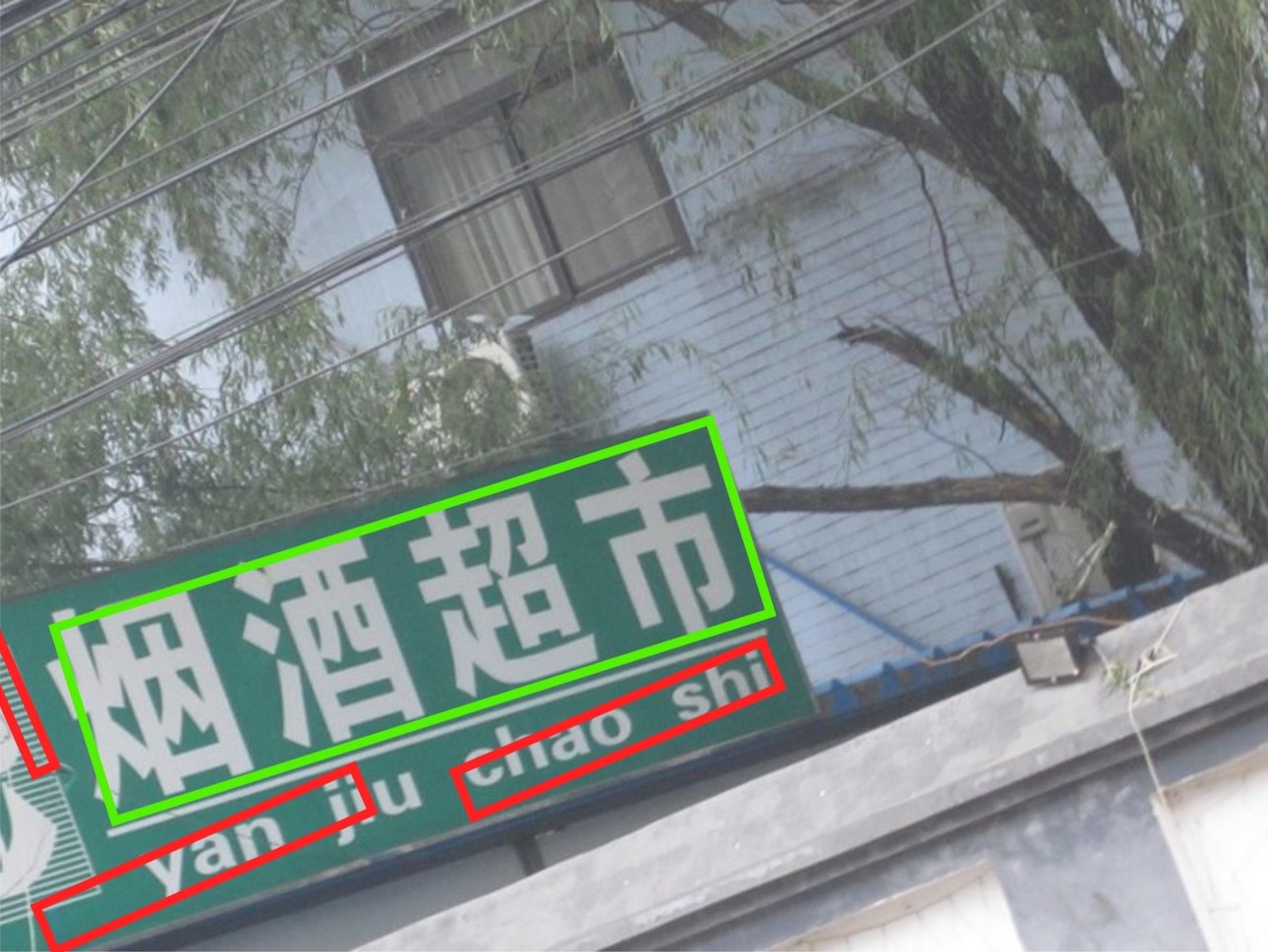}
\caption{}
\label{fig:aug_show_det}
\end{subfigure}
\begin{subfigure}[t]{0.19\textwidth}
\includegraphics[width=\textwidth]{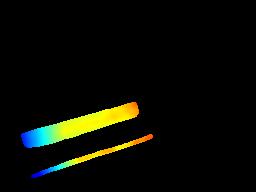}
\caption{}
\label{fig:aug_show_geo_aug}
\end{subfigure}
\begin{subfigure}[t]{0.19\textwidth}
\includegraphics[width=\textwidth]{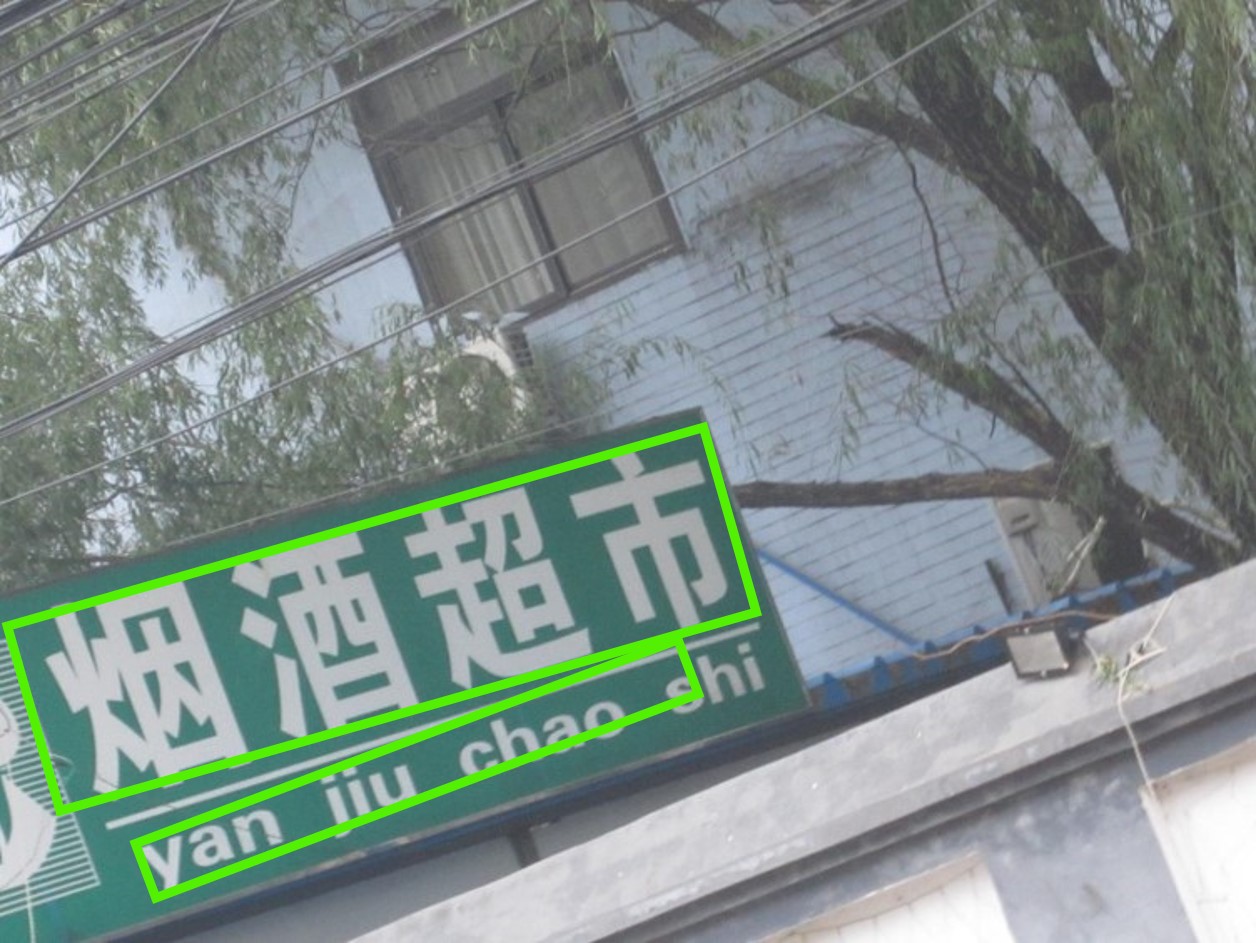}
\caption{}
\label{fig:aug_show_det_aug}
\end{subfigure}
\caption{The inclusion of augmented images improves the scene text detection: With the inclusion of the augmented images in training, more consistent text feature maps and more complete scene text detections are produced as shown in (d) and (e), as compared with those produced by the baseline model (trained using original training images only) shown in (b) and (c). The coloring in the text feature maps shows the distance information predicted by regressor (blue denotes short distances and red denotes long distance).}
\label{fig:aug_show}
\end{figure}

The proposed bootstrapping based scene text image augmentation technique improves the consistency of predicted text feature map as well as the performance of regression greatly as illustrated in Fig. \ref{fig:aug_show}. For the sample images in Fig. \ref{fig:aug_show_orig}, Figs. \ref{fig:aug_show_geo} and \ref{fig:aug_show_geo_aug} show the text feature maps that are produced by the baseline model (trained by using the original training images) and the augmented model (trained by further including the augmented sample images), respectively (training details to be described in Section \ref{sec:text_det}). The coloring in the text feature maps shows the distance to the left-side boundary as predicted by regressor - blue denotes short distance and red denotes long distance. Figs. \ref{fig:aug_show_det} and \ref{fig:aug_show_det_aug} show the corresponding detection boxes, respectively, where red boxes show false detections and green boxes show correct detections. It can be seen that the inclusion of the augmented images helps to produce more consistent text feature maps as well as smoother geometrical distance maps (for regression of text boxes) which leads to more complete instead of broken scene text detections.

\subsection{Semantics-Aware Text Borders} \label{sec:border}
\begin{figure}[t]
\centering
\includegraphics[width=\textwidth]{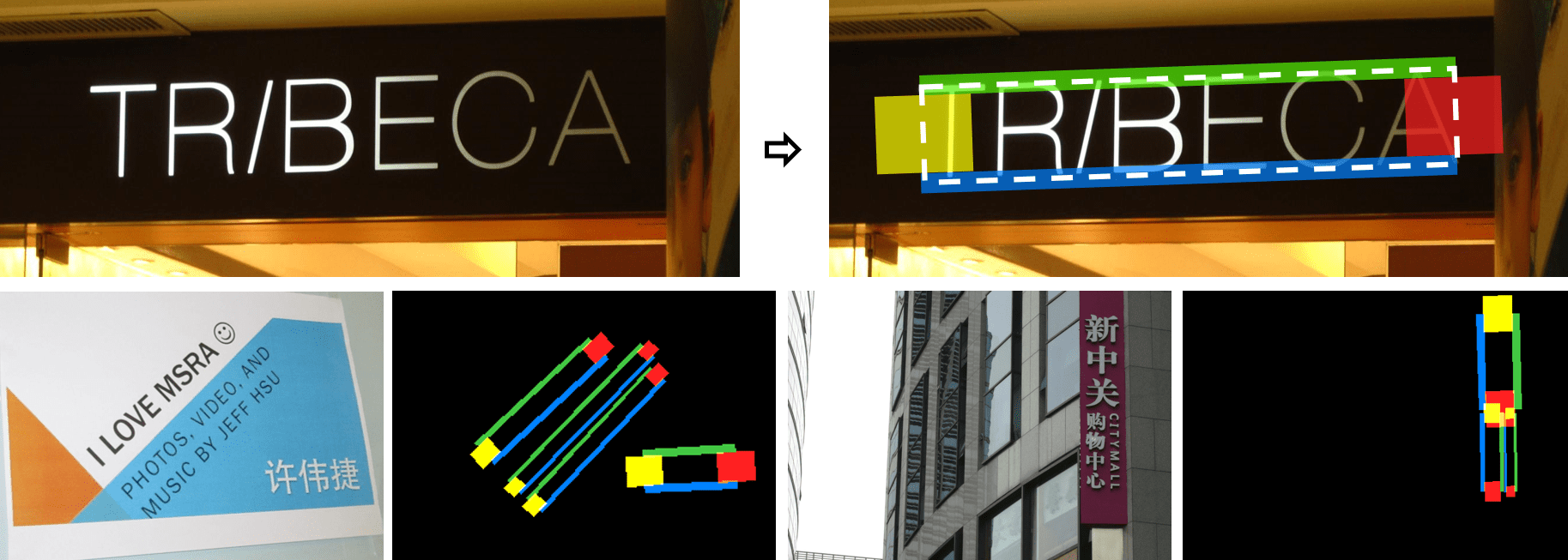}
\caption{Semantics-aware text border detection: Four text border segments are automatically extracted for each text annotation box including a pair of short-edge text border segments in yellow and red colors and a pair of long-edge text border segments in green and blue colors. The four types of text border segments are treated as four types of objects and used to train deep network models, and the trained model is capable of detecting the four types of text border segments as illustrated in Fig. \ref{fig:border_show_border}.}
\label{fig:labeling}
\end{figure}

We extract two pairs of semantics-aware text border segments for each scene text annotation as illustrated in Fig. \ref{fig:labeling}. With $W$ and $L$ denoting the width and length of a text annotation box (TAB), a pair of long text border segments in green and blue colors can be extracted along the two long edges of the TAB as illustrated in Fig. \ref{fig:labeling}, where the segment length is set at $L$ and the segment width is empirically set at $0.2*W$. In addition, the center line of the two long text border segments overlaps perfectly with the long edges of TAB so that the text border segments capture the transition from text to background or vice versa. 

A pair of short text border segments can also be extracted based on the TAB as illustrated in Fig. \ref{fig:labeling}. In particular, the dimension along the TAB width is set at $0.8*W$ which fits perfectly in between the two long text segments. Another dimension along the TAB length is set the same as $W$ with which the trained text border detector can detect a certain amount of text pixels to be used in text bounding box regression. Similarly, the center line of the short text border segments (along the TAB width) overlaps perfectly with the TAB short edge so that the extracted text border segments capture the transition from text to background or vice versa. 

\begin{figure}[t]
\centering
\begin{subfigure}[t]{0.24\textwidth}
\includegraphics[width=\textwidth]{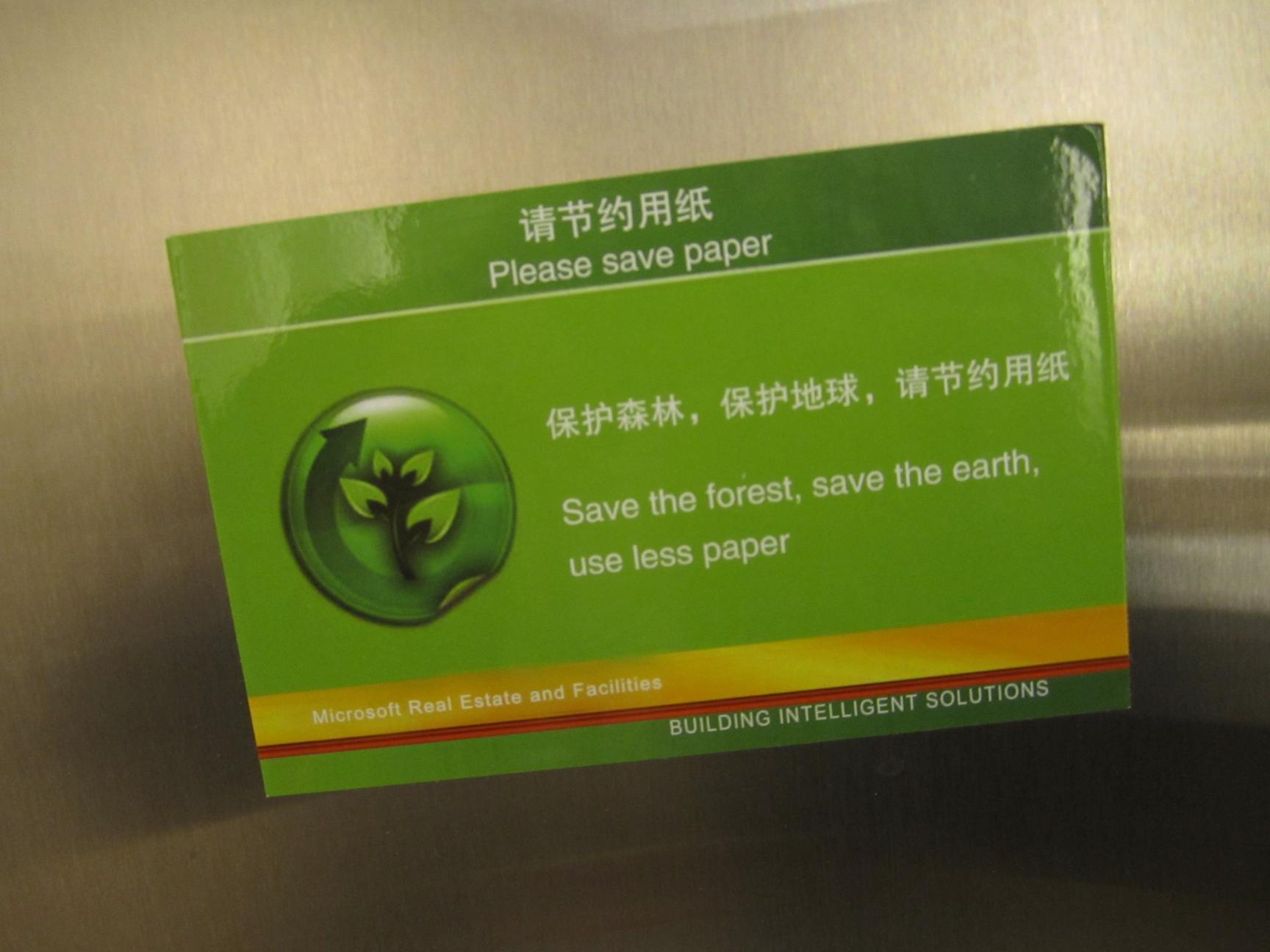}
\end{subfigure}
\begin{subfigure}[t]{0.24\textwidth}
\includegraphics[width=\textwidth]{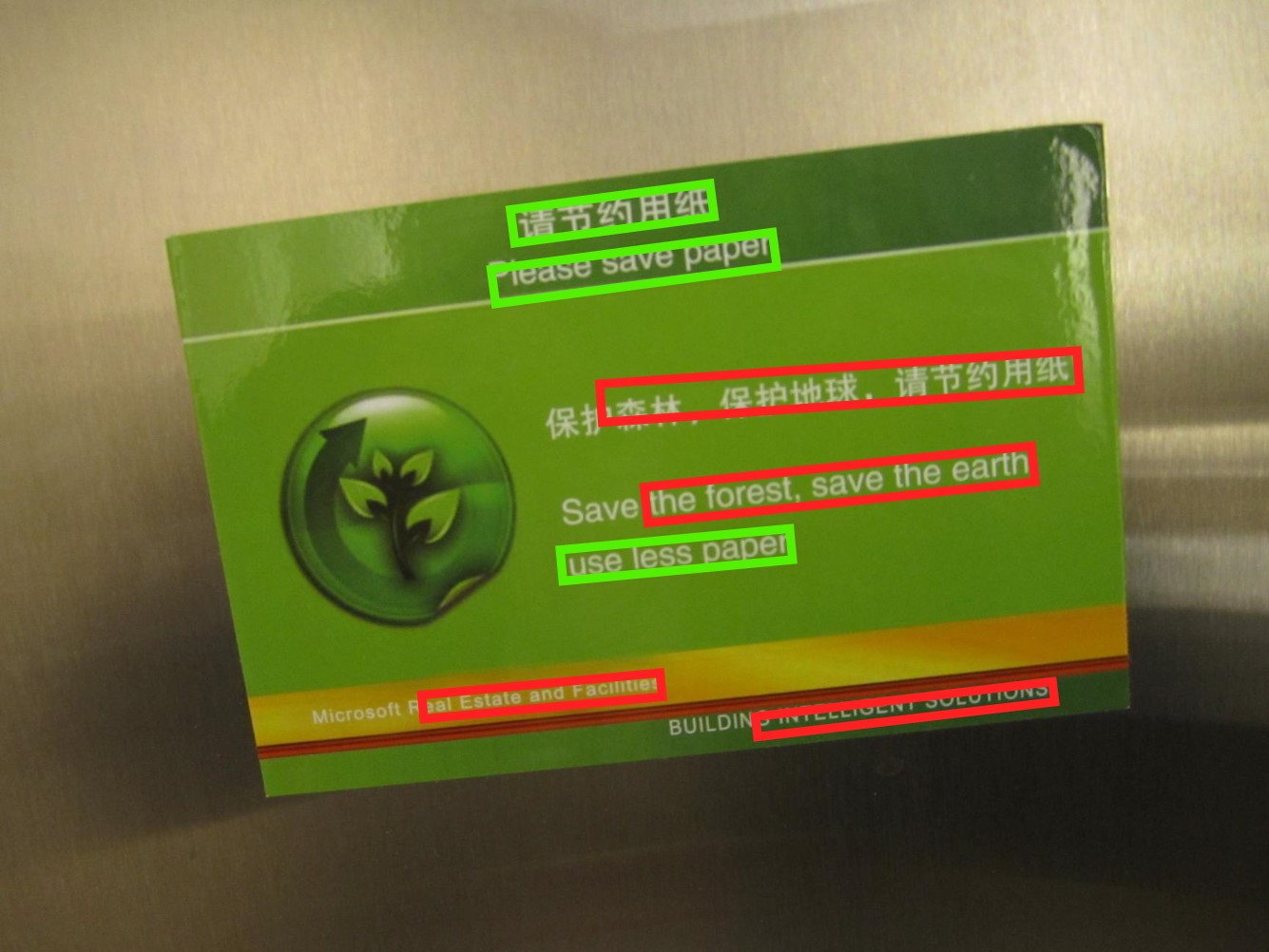}
\end{subfigure}
\begin{subfigure}[t]{0.24\textwidth}
\includegraphics[width=\textwidth]{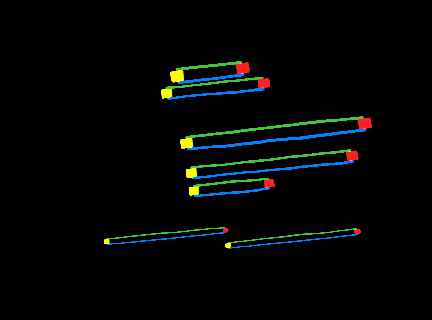}
\end{subfigure}
\begin{subfigure}[t]{0.24\textwidth}
\includegraphics[width=\textwidth]{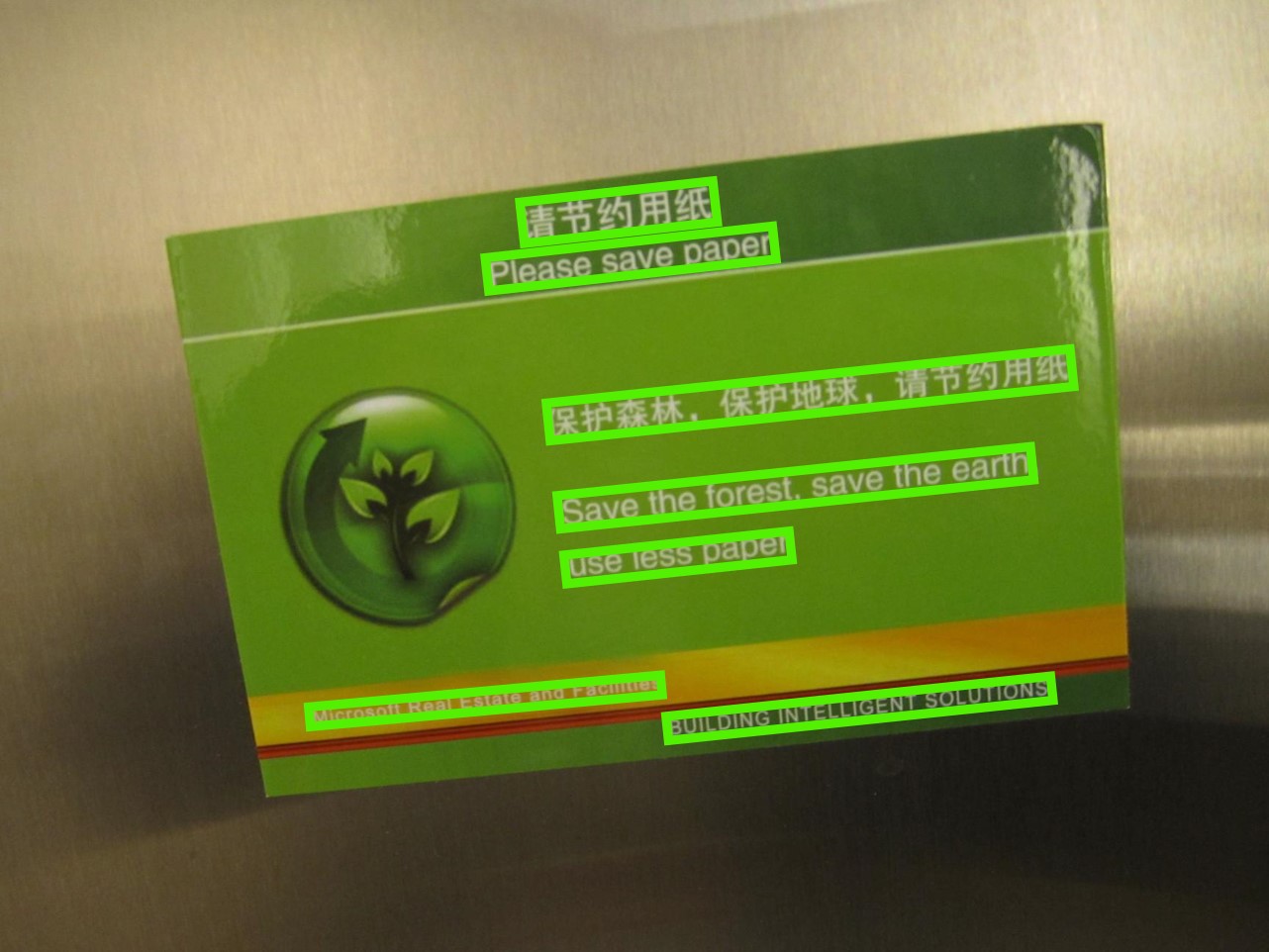}
\end{subfigure}
\begin{subfigure}[t]{0.24\textwidth}
\includegraphics[width=\textwidth]{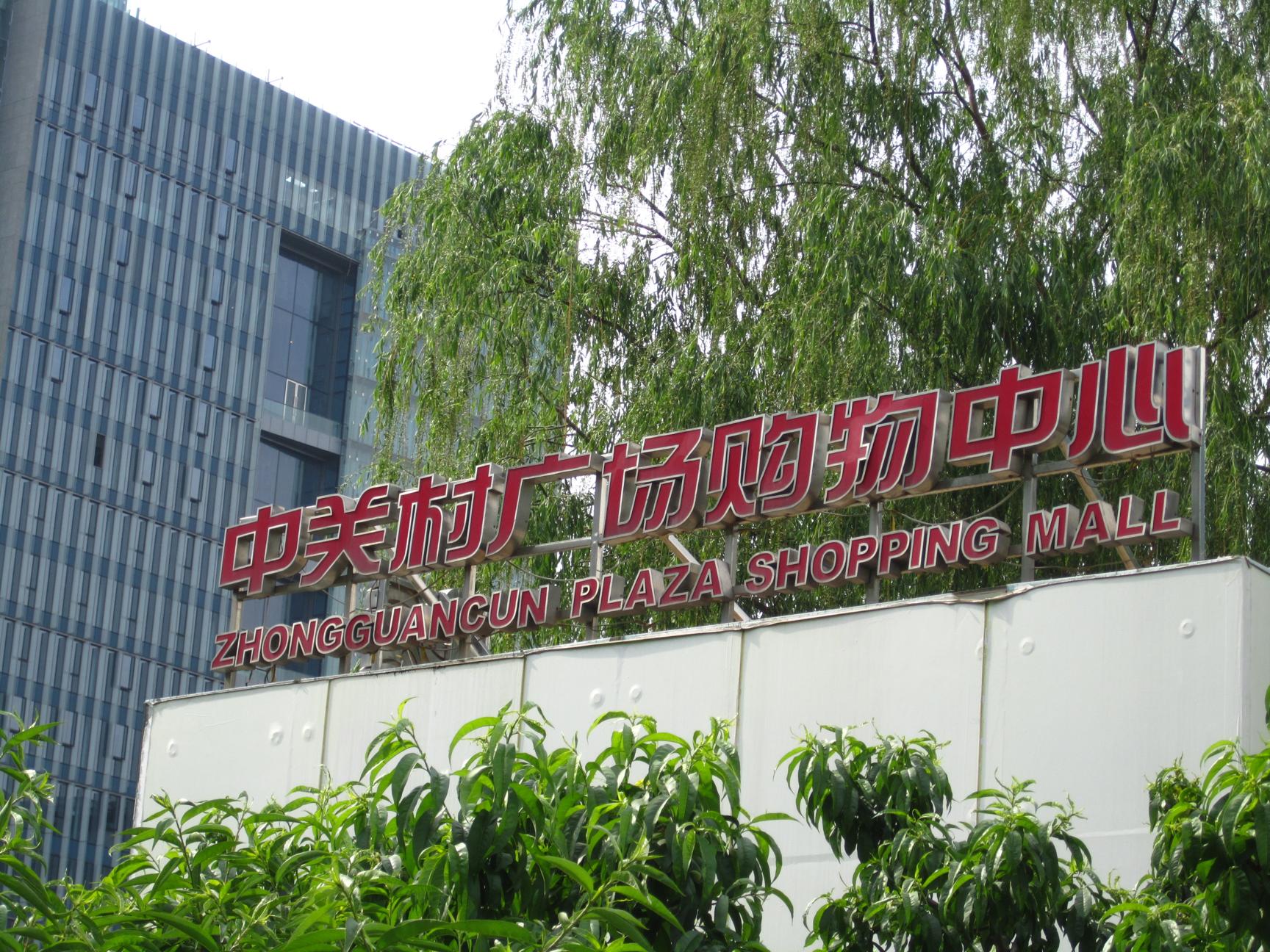}
\end{subfigure}
\begin{subfigure}[t]{0.24\textwidth}
\includegraphics[width=\textwidth]{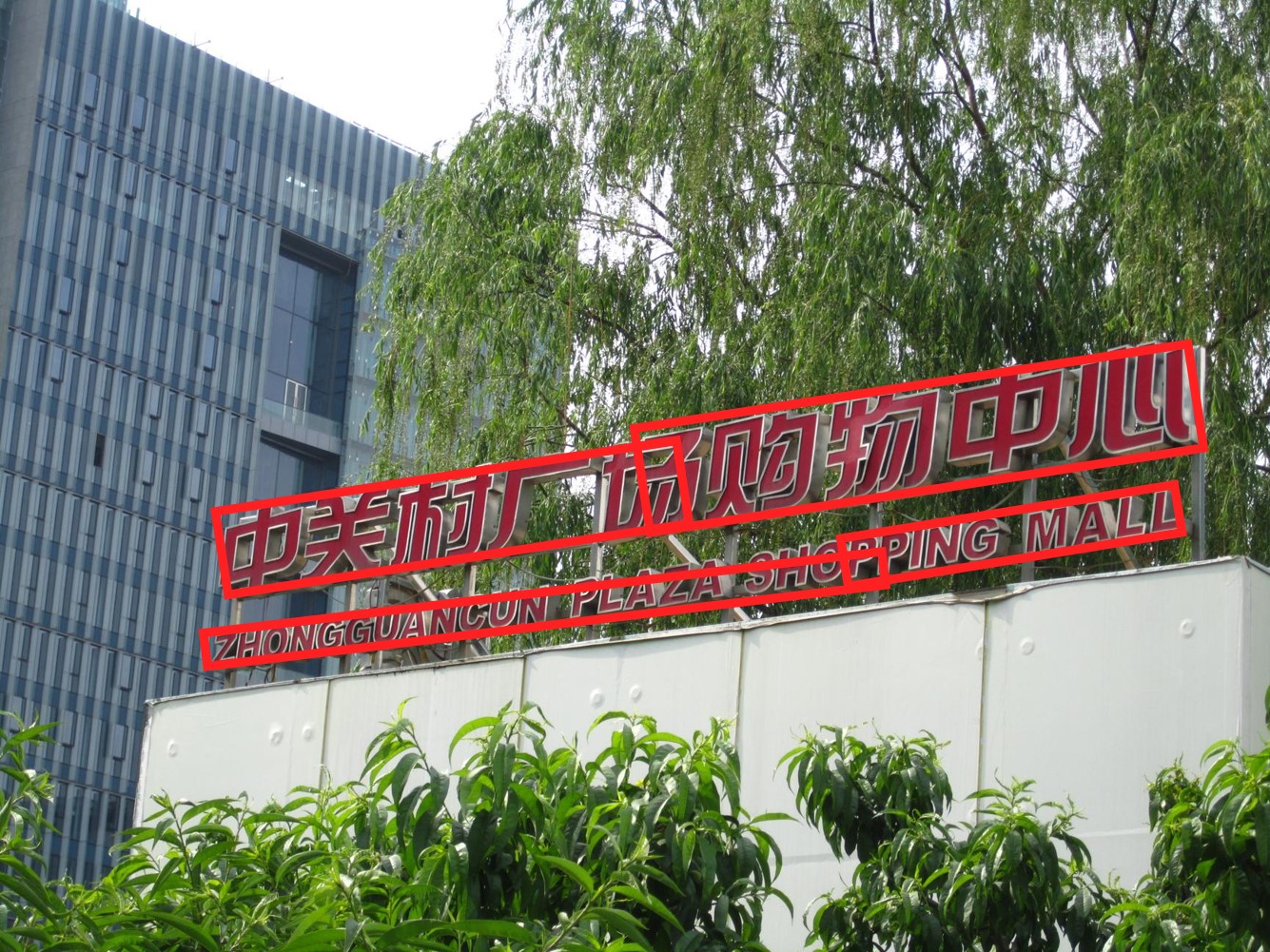}
\end{subfigure}
\begin{subfigure}[t]{0.24\textwidth}
\includegraphics[width=\textwidth]{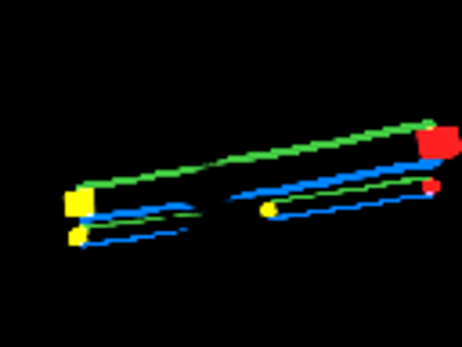}
\end{subfigure}
\begin{subfigure}[t]{0.24\textwidth}
\includegraphics[width=\textwidth]{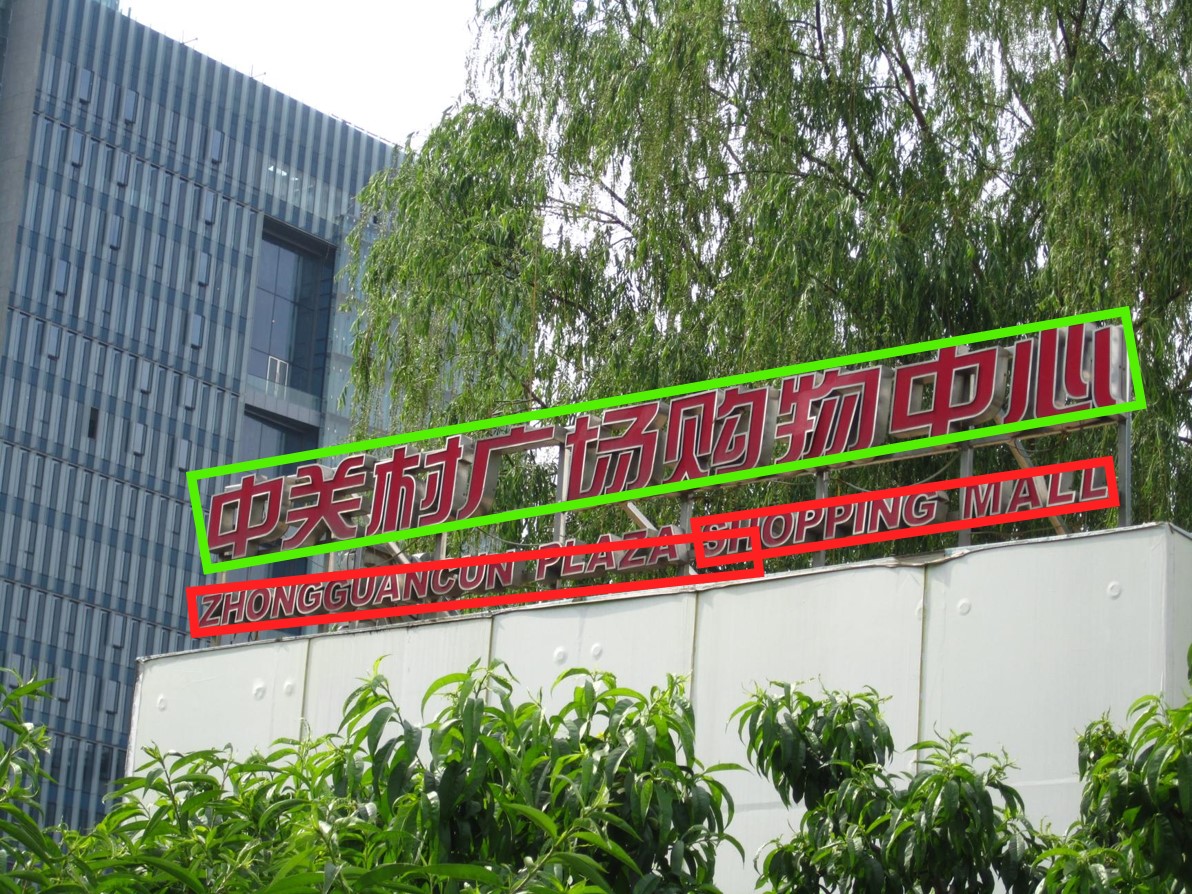}
\end{subfigure}
\begin{subfigure}[t]{0.24\textwidth}
\includegraphics[width=\textwidth]{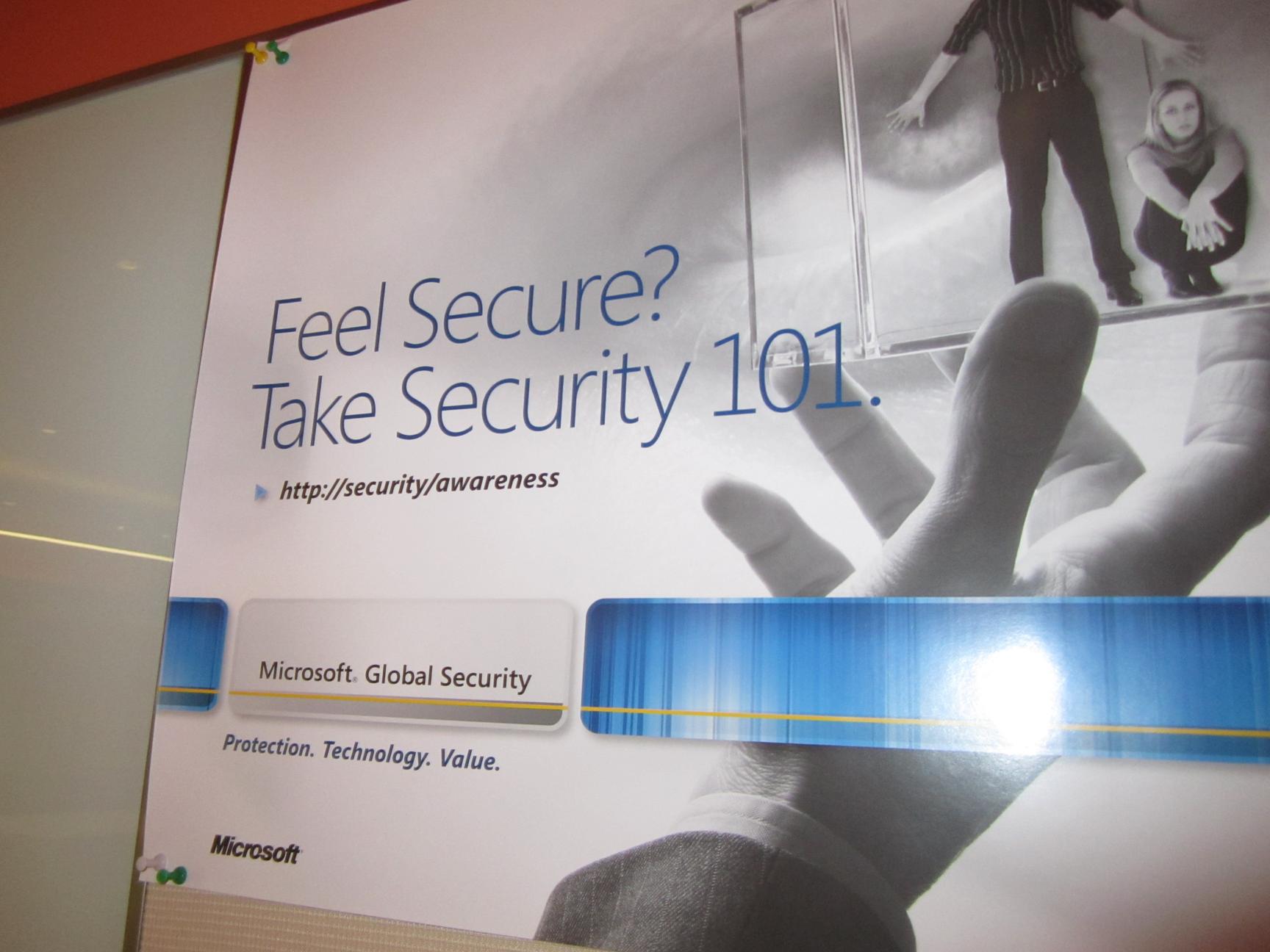}
\caption{}
\label{fig:border_show_orig}
\end{subfigure}
\begin{subfigure}[t]{0.24\textwidth}
\includegraphics[width=\textwidth]{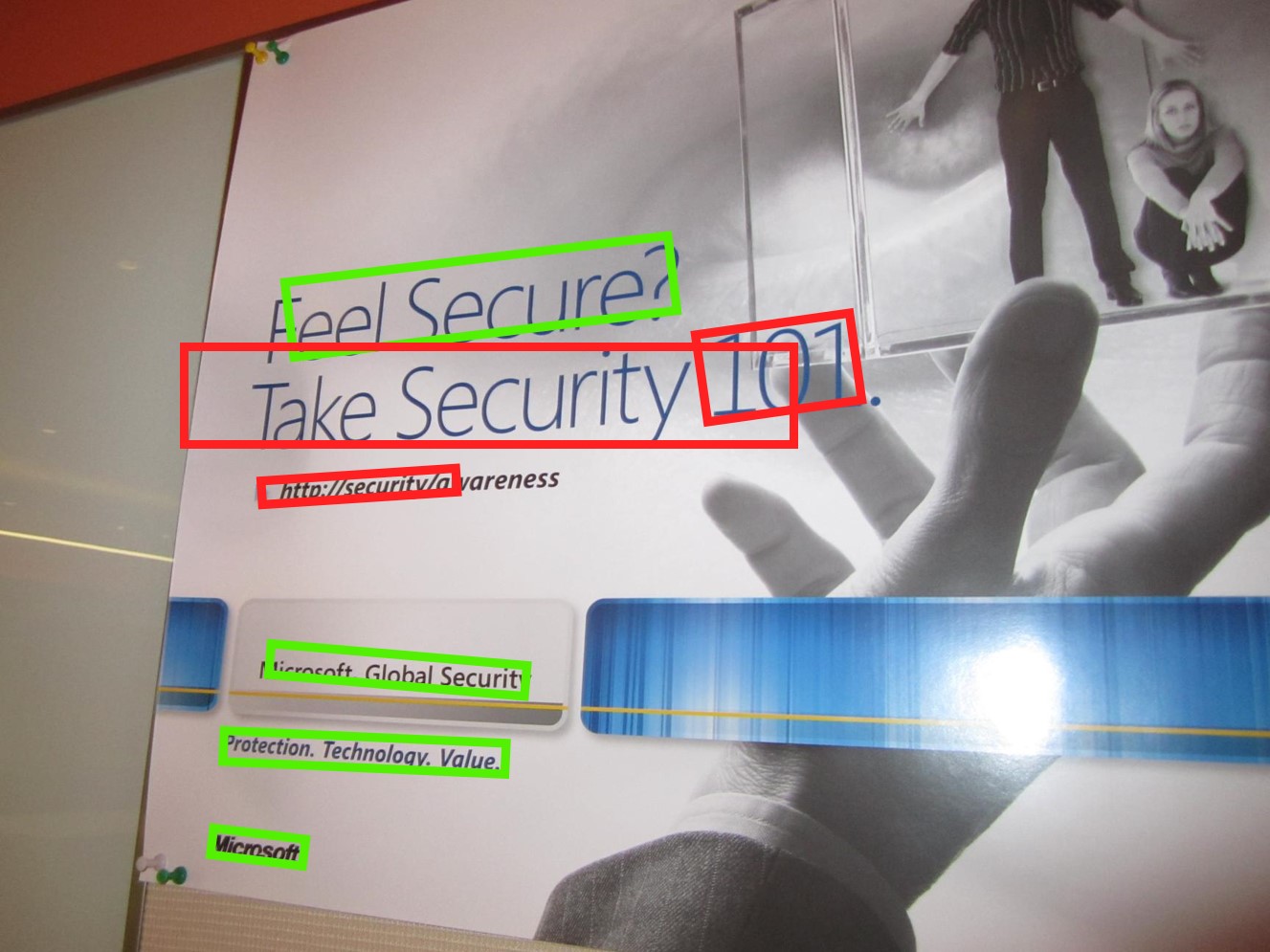}
\caption{}
\label{fig:border_show_det}
\end{subfigure}
\begin{subfigure}[t]{0.24\textwidth}
\includegraphics[width=\textwidth]{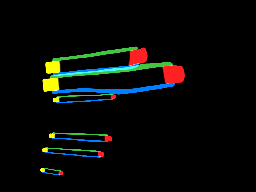}
\caption{}
\label{fig:border_show_border}
\end{subfigure}
\begin{subfigure}[t]{0.24\textwidth}
\includegraphics[width=\textwidth]{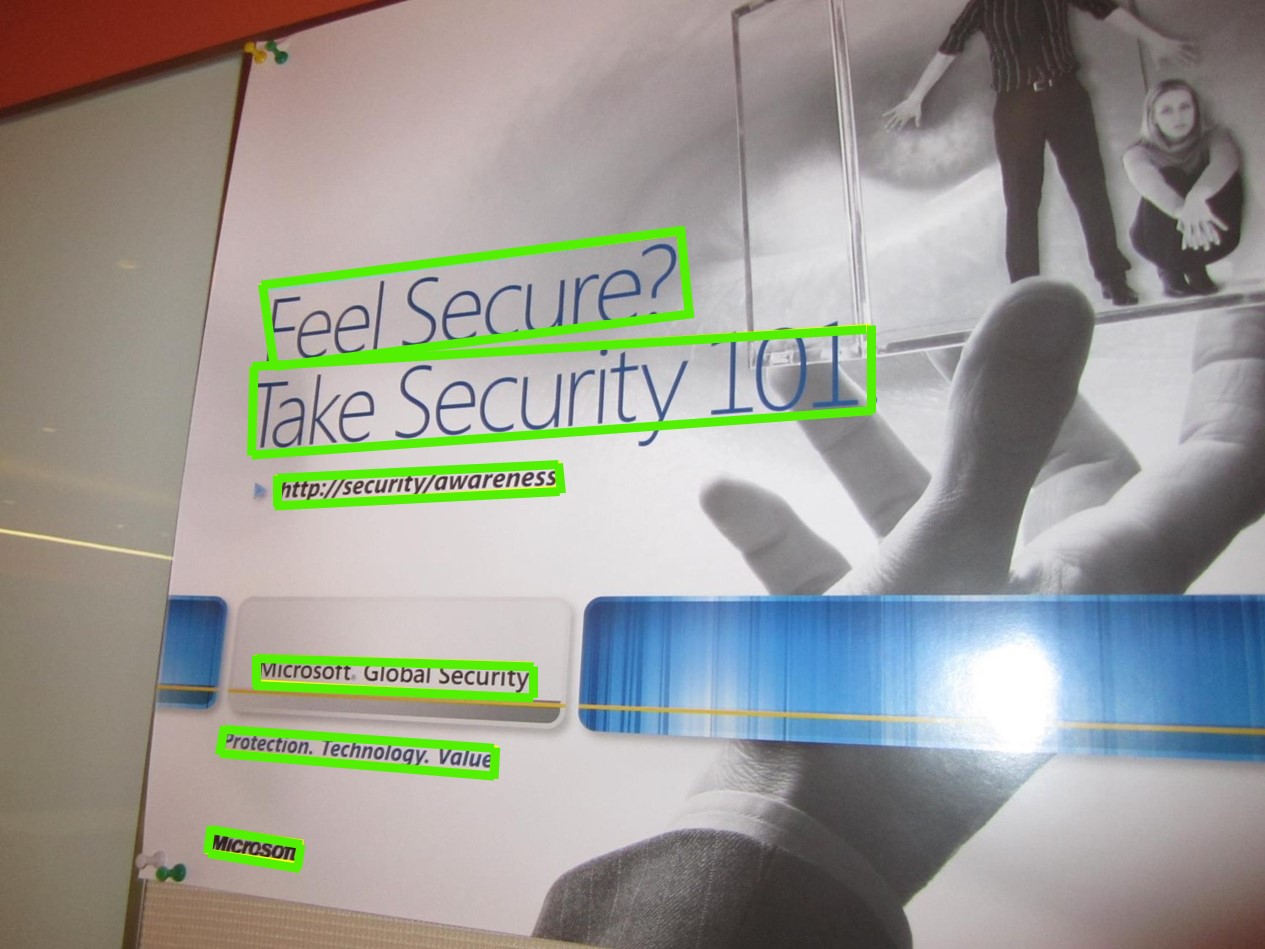}
\caption{}
\label{fig:border_show_det_border}
\end{subfigure}
\caption{The use of semantics-aware text borders improves scene text detection: With the identified text border semantics information as illustrated in (c), scene texts can be localized much more accurately as illustrated in (d) as compared with the detections without using the border semantics information as illustrated in (b). Green boxes give correct detections and red boxes give false detections.}
\label{fig:border_show}
\end{figure}

The use of the semantics-aware text borders helps to improve the localization accuracy of the trained scene text detection model greatly (training details to be described in Sec. \ref{sec:text_det}). With the identified text border semantics as shown in Fig. \ref{fig:border_show_border} (four colors are for illustration only), text pixels around the text line ends can be determined by the overlap between the short text border segments and the predicted text feature map. The text bounding box can thus be regressed by using the text pixels lying around the text line ends which often leads to accurate text localization as illustrated in Fig. \ref{fig:border_show_det_border}. The reason is that text pixels around the middle of texts are far from the text box vertices for long words or text lines which can easily introduce regression errors and lead to inaccurate localization as illustrated in Fig. \ref{fig:border_show_det}. At the other end, the long text border segments also help for better scene text detection performance. In particular, the long text border segments can be exploited to separate text lines when neighboring text lines are close to each other.

\subsection{Scene Text Detection} \label{sec:text_det}
The original scene text annotations, together with the augmented images and the extracted semantics-aware text borders as described in Sect. \ref{sec:aug} and \ref{sec:border}, are fed to a multi-channel FCN to train a scene text detection model. The training aims to minimize the following multi-task loss function:

\begin{equation}
\mathcal{L}=\mathcal{L}_{cls}+\lambda_{loc}*\mathcal{L}_{loc}+\lambda_{brd}*\mathcal{L}_{brd}
\end{equation}
where $\mathcal{L}_{cls}$, $\mathcal{L}_{loc}$ and $\mathcal{L}_{brd}$ refer to loss of text feature (confidence score of each pixel being a text pixel), regression (distances from each pixel to four sides of text boundaries) and border feature (confidence score of each pixel being a border pixel), respectively. Parameters $\lambda_{loc}$ and $\lambda _{brd}$ are weights of the corresponding losses which are empirically set at 1.0 in our system.

For the regression loss $\mathcal{L}_{loc}$, we adopt the IoU loss \cite{yu2016unitbox} in training. For the classification losses $\mathcal{L}_{cls}$ and $\mathcal{L}_{brd}$, we use Dice’s Coefficient \cite{milletari2016v} which is a widely-used in image segmentation tasks. Given a ground truth region \textit{G} and a predicted region \textit{P}, the Dice’s Coefficient is defined by:

\begin{equation}
\mathcal{L}_{brd}= \frac{2*|G\cap P|}{|G|+|P|}
\end{equation}

\begin{algorithm}[t]
\caption{Text bounding box detection.} \label{detector}
\begin{algorithmic}[1]
\STATE $\textbf{Input}$: Regressor $R$, text region map $T$ and text border region map $B$
\STATE $\textbf{Output}$: A list of text bounding boxes $\textbf{BB}$
\STATE Delineate text lines $t$ in $T$ using the long text border regions in $B$
\STATE Determine left-side and right-side regressing text pixels $p_l$ and $p_r$ by overlaps between the delineated $t$ and the two short text border regions in $B$
\STATE Derive two sets of text boxes $\textbf{BB}_{l}$ and $\textbf{BB}_{r}$ by regressing $p_l$ and $p_r$
\STATE $\textbf{BB}$ $\leftarrow$ $\Phi$
\FOR{each box in $\textbf{BB}_{l}$ and $\textbf{BB}_{r}$} 
\IF{the two boxes are regressed from text pixels of the same $t$} 
\STATE Merge the two boxes and add the merged box to $\textbf{BB}$
\ENDIF 
\ENDFOR
\STATE Apply NMS to $\textbf{BB}$
\end{algorithmic} 
\end{algorithm}

Given a test image, our trained scene text detector produces three maps including a text feature map, a text border feature map, and a regressor. The text border feature map has four channels which give a pair of short text border segments and a pair of long text border segments as illustrated in Fig. \ref{fig:border_show_border}. The regressor also has four channels that predict the distances to the upper, lower, left and right text boundaries, respectively, as illustrated in Figs. \ref{fig:aug_show_det} and \ref{fig:aug_show_det_aug} (which shows one channel distance to the left-side text boundary).

Algorithm \ref{detector} shows how the text bounding boxes are derived from the outputs of the trained scene text detector. Given a text feature map and a text border feature map, a text region map and four text border region maps are first determined (as the algorithm inputs) by global thresholding where the threshold is simply estimated by the mean of the respective feature map. Overlaps between the text region map and four text border region maps can then be determined. Text lines can thus be delineated by removing the overlaps between the text region map and the two long text border region maps. Further, text bounding box vertices at the left and right text line ends can be predicted by regressing the text pixels that overlap with the left-side and right-side text border region maps, respectively. Finally, the text bounding box is determined by merging the regressed left-side and right-side text box vertices.

\begin{table}[t]
\centering
\caption{Recall (R), precision (P) and f-score (F) of different scene text detection methods over the MSRA-TD500 and ICDAR2013 datasets.}\label{tab:msra_icdar13}
\begin{tabular}{|c|c|c|c|c|c|c|c|c|}
\hline
\multicolumn{4}{|c|}{MSRA-TD500} && \multicolumn{4}{|c|}{ICDAR2013}\\
\hline
Method & R & P & F&& Method & R & P & F \\
\hline
\hline
Kang \textit{et al.} \cite{kang2014orientation} & 62.0 & 71.0 & 66.0 && He \textit{et al.} \cite{He2017Deep} & 81.0 & 92.0 & 86.0\\
\hline
Yin \textit{et al.} \cite{yin2015multi} & 63.0 & 81.0 & 71.0 && Tian \textit{et al.} \cite{Tian_2017_ICCV} & 83.1 & 91.1 & 86.9\\
\hline
Zhang \textit{et al.} \cite{zhang2016multi} & 67.0 & 83.0 & 74.0 && He \textit{et al.} \cite{He2017Single} & 86.0 & 88.0 & 87.0\\
\hline
He \textit{et al.} \cite{He2017Deep} & 70.0 & 77.0 & 74.0 && Zhou \textit{et al.} \cite{Zhou_2017_CVPR} & 82.7 & 92.6 & 87.7\\
\hline
Yao \textit{et al.} \cite{yao2016scene} & 75.3 & 76.5 & 75.9 && Jiang \textit{et al.} \cite{jiang2017r2cnn} & 82.6 & \textbf{93.6} & 87.7\\
\hline
Zhou \textit{et al.} \cite{Zhou_2017_CVPR} & 67.4 & \textbf{87.3} & 76.1 && He \textit{et al.} \cite{he2018end} & 83.0 & 93.0 & 88.0\\
\hline
Shi \textit{et al.} \cite{Shi_2017_CVPR} & 70.0 & 86.0 & 77.0 && Tian \textit{et al.} \cite{tian2016detecting} & 87.0 & 88.0 & 88.0\\
\hline
Wu \textit{et al.} \cite{Wu_2017_ICCV} & \textbf{78.0} & 77.0 & 77.0 && Hu \textit{et al.} \cite{Hu2017ICCVWordSup} & \textbf{87.5} & 93.3 & \textbf{90.3}\\
\hline\hline
Baseline(ResNet) & 73.4 & 70.3 & 71.8 && Baseline(ResNet) & 79.3 & 86.9 & 83.0\\ 
\hline
Border(ResNet) & 72.0 & 76.4 & 74.3 && Border(ResNet) & 84.5 & 85.4 & 84.9\\
\hline
Aug.(ResNet) & 71.1 & 77.7 & 74.3 && Aug.(ResNet) & 86.7 & 83.8 & 85.2\\
\hline
Aug.+Border(ResNet) & 73.3 & 80.7 & 76.8 && Aug.+Border(ResNet) & 86.9 & 87.8 & 87.4\\
\hline
Aug.+Border(DenseNet) & 77.4 & 83.0 & \textbf{80.1} && Aug.+Border(DenseNet) & 87.1 & 91.5 & 89.2\\
\hline
\end{tabular}
\end{table}

\section{Experiments} \label{sec:experiment}
\subsection{Datasets and Evaluation Metrics} \label{sec:dataset}
\textbf{MSRA-TD500}\footnote{http://tc11.cvc.uab.es/datasets/MSRA-TD500\_1}\cite{yao2012detecting} comprise 300 training images and 200 testing images with scene texts printed in either Chinese or English. For each training image, annotations at either word or text line level is provided, where each annotation consists of a rectangle box and the corresponding box rotation angle. Due to the very small number of training images, 400 training images in the HUST-TR400 \footnote{http://mclab.eic.hust.edu.cn/UpLoadFiles/dataset/HUST-TR400.zip} \cite{yao2014unified} are included in training.

\textbf{ICDAR2013}\footnote{http://rrc.cvc.uab.es/?ch=2\&com=introduction}\cite{karatzas2013icdar} consists of 229 training images and 233 testing images with texts in English. The text annotations are at word level, and no rotation angles are provided as most captured scene texts are almost horizontal. We also include training images from ICDAR2015 in training.

\textbf{ICDAR2017-RCTW}\footnote{http://www.icdar2017chinese.site:5080/dataset/}\cite{shi2017icdar2017} comprises 8,034 training images and 4,229 testing images with scene texts printed in either Chinese or English. The images are captured from different sources including street views, posters, screen-shot, etc. Multi-oriented words and text lines are annotated using quadrilaterals.

\textbf{ICDAR2017-MLT}\footnote{http://rrc.cvc.uab.es/?ch=8}\cite{nayef2017icdar2017} comprise 7,200 training images, 1,800 validation images and 9,000 testing images with texts printed in 9 languages including Chinese, Japanese, Korean, English, French, Arabic, Italian, German and Indian. Most annotations are at word level while texts in non-Latin languages like Chinese are annotated at text-line level. Similar to ICDAR2017-RCTW, texts in this dataset are also multi-oriented with text annotated using quadrilaterals.

\subsubsection{Evaluation Metrics}
For MSRA-TD500, we use the evaluation protocol in \cite{wang2011end}. For ICDAR2013, ICDAR2017-RCTW and ICDAR2017-MLT, we perform evaluations by using the online evaluation systems that are provided by the respective dataset creators. In particular, one-to-many (one rectangle corresponds to many rectangles) and many-to-one (many rectangles correspond to one rectangle) matches are adopted for better evaluation for the ICDAR2013 dataset.

\subsection{Implementation Details} \label{sec:implement}
The network is optimized by Adam \cite{kingma2014adam} optimizer with starting learning rate of $10^{-4}$ and batch size of 16. Images are randomly resized with ratio of 0.5, 1, 2, or 3 and cropped into 512 x 512 without crossing texts before training. 20 augmented images are sampled for each training images by using the proposed data augmentation technique. The whole experiments are conducted on Nvidia DGX-1. All our models are fine-tuned from a pre-trained model using the ImageNet dataset \cite{deng2009imagenet}. Two base networks including ResNet \cite{he2016deep} and DenseNet \cite{huang2016densely} are implemented for evaluation. Multi-scale evaluation is implemented by resizing the longer side of test images to 256, 512, 1024, 2048 pixels. 

\begin{table}[t]
\centering
\caption{ Recall (R), precision (P) and f-score (F) of different detection methods over the ICDAR2017-RCTW and ICDAR2017-MLT datasets.}\label{tab:icdar17_mlt}
\begin{tabular}{|c|c|c|c|c|c|c|c|c|}
\hline
\multicolumn{4}{|c|}{ICDAR2017-RCTW} && \multicolumn{4}{|c|}{ICDAR2017-MLT}\\
\hline
Method & R & P & F && Method & R & P & F\\
\hline\hline
gmh \cite{shi2017icdar2017} & 57.8 & 70.6 & 63.6 && Sensetime OCR \cite{nayef2017icdar2017} & \textbf{69.4} & 56.9 & 62.6\\ 
\hline
NLPR\_PAL \cite{shi2017icdar2017} & 57.3 & 77.2 & 65.8 && SCUT\_DLVClab \cite{nayef2017icdar2017} & 54.5 & \textbf{80.3} & 65.0\\ 
\hline
Foo\&Bar \cite{shi2017icdar2017} & \textbf{59.5} & 74.4 & 66.1 && NLPR\_PAL \cite{He2017Deep} & 57.9 & 76.7 & 66.0\\ 
\hline\hline
Baseline(ResNet) & 52.2 & 66.6 & 58.5 &&Baseline(ResNet) & 60.9 & 64.5 & 62.6\\ 
\hline
Border(ResNet) & 58.5 & 74.2 & 65.4 && Border(ResNet) & 60.6 & 73.9 & 66.6\\
\hline
Border(DenseNet) & 58.8 & \textbf{78.2} & \textbf{67.1} && Border(DenseNet) & 62.1 & 77.7 & \textbf{69.0}\\
\hline
\end{tabular}
\end{table}

\subsection{Experimental Results} \label{sec:experiment}
\textbf{Quantitative Results.} Table \ref{tab:msra_icdar13} shows quantitative experimental results on the MSRA-TD500 and ICDAR2013 datasets as well as comparisons with state-of-the-art methods. As Table \ref{tab:msra_icdar13} shows, five models are trained including: 1) `Baseline (ResNet)' that is trained by using ResNet-50 and the original training images as described in Sec. \ref{sec:dataset}, 2) `Border (ResNet)' that is trained by including text border segments as described in Sec. \ref{sec:border}, 3) `Aug. (ResNet)' that is trained by including augmented scene text images as described in Sec. \ref{sec:aug}, 4) `Aug.+Border (ResNet)' that is trained by including both text border segments and augmented images, and 5) `Aug.+Border (DenseNet)' that is trained by using DenseNet-121 with the same training data as the `Aug.+Border (ResNet)'.

As Table \ref{tab:msra_icdar13} shows, the detection models using either semantics-aware text borders or augmented images or the both outperform the baseline model consistently. In addition, the models using both text borders and augmented images outperform the ones using either text borders or augmented images alone. Further, the trained models outperform the state-of-the-art methods clearly when the DenseNet-121 is used, demonstrating the superior performance of the proposed technique. We observe that the performance improvement is mainly from higher precisions for the MSRA-TD500 dataset as compared with higher recalls for the ICDAR2013 dataset. This inconsistency is largely due to the different evaluation methods for the two datasets, i.e. the evaluation of the MSRA-TD500 follows one-to-one (one rectangle corresponds to one rectangles) match whereas the evaluation of the ICDAR2013 follows one-to-many and many-to-one. As studied in \cite{Wu_2017_ICCV}, the ICDAR2013 online evaluation system usually produces lower precision as compared with the real values. We conjecture that the actual precision by our method should be higher than what is presented in Table 1.

\begin{figure}[t]
\centering
\begin{subfigure}[t]{0.24\textwidth}
\includegraphics[width=\textwidth]{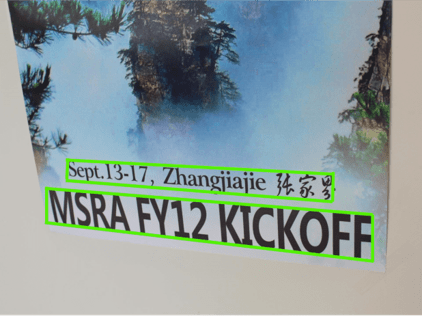}
\end{subfigure}
\begin{subfigure}[t]{0.24\textwidth}
\includegraphics[width=\textwidth]{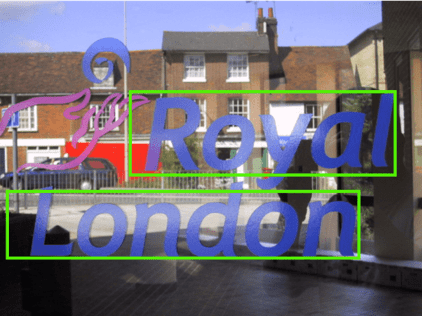}
\end{subfigure}
\begin{subfigure}[t]{0.24\textwidth}
\includegraphics[width=\textwidth]{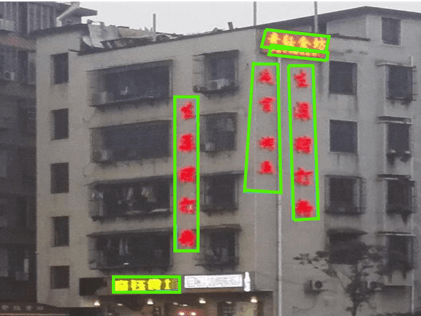}
\end{subfigure}
\begin{subfigure}[t]{0.24\textwidth}
\includegraphics[width=\textwidth]{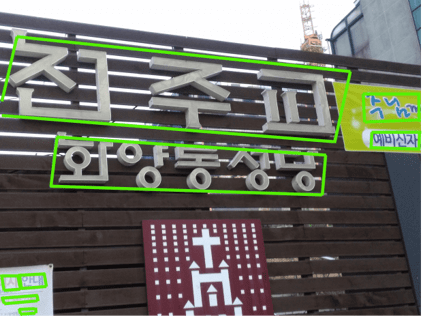}
\end{subfigure}
\begin{subfigure}[t]{0.24\textwidth}
\includegraphics[width=\textwidth]{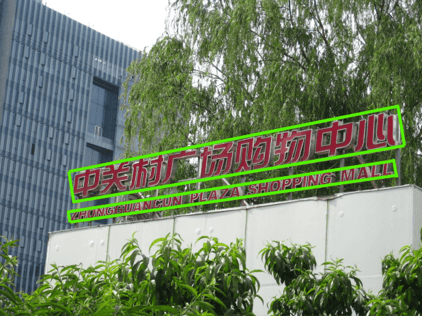}
\end{subfigure}
\begin{subfigure}[t]{0.24\textwidth}
\includegraphics[width=\textwidth]{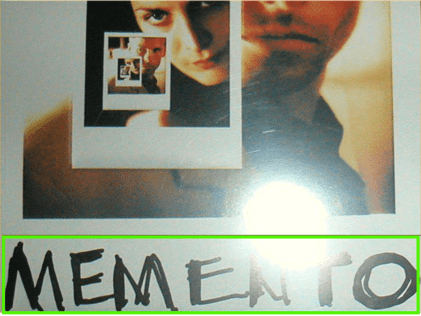}
\end{subfigure}
\begin{subfigure}[t]{0.24\textwidth}
\includegraphics[width=\textwidth]{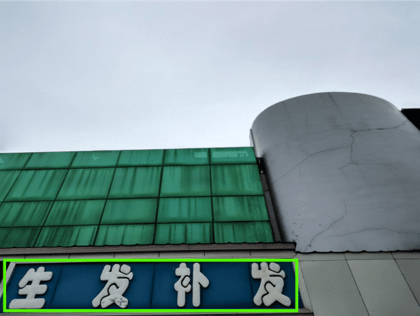}
\end{subfigure}
\begin{subfigure}[t]{0.24\textwidth}
\includegraphics[width=\textwidth]{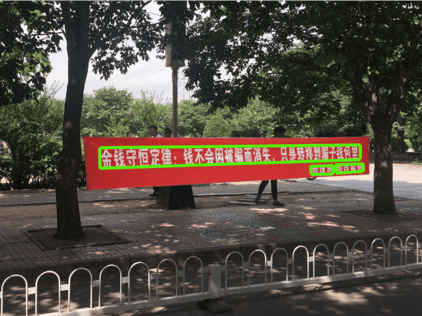}
\end{subfigure}
\begin{subfigure}[t]{0.24\textwidth}
\includegraphics[width=\textwidth]{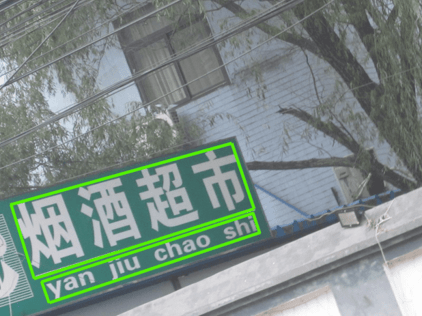}
\caption{}
\end{subfigure}
\begin{subfigure}[t]{0.24\textwidth}
\includegraphics[width=\textwidth]{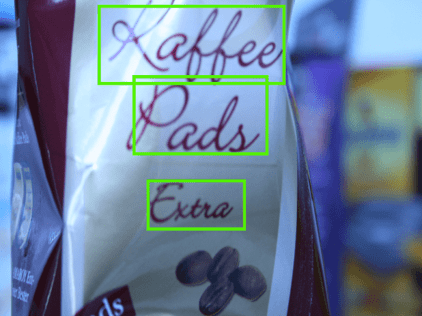}
\caption{}
\end{subfigure}
\begin{subfigure}[t]{0.24\textwidth}
\includegraphics[width=\textwidth]{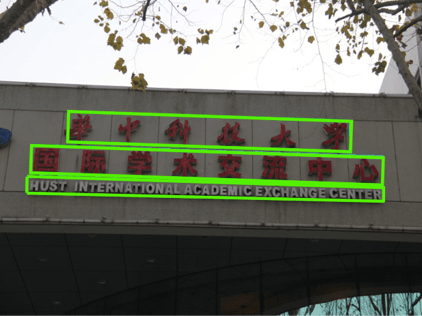}
\caption{}
\end{subfigure}
\begin{subfigure}[t]{0.24\textwidth}
\includegraphics[width=\textwidth]{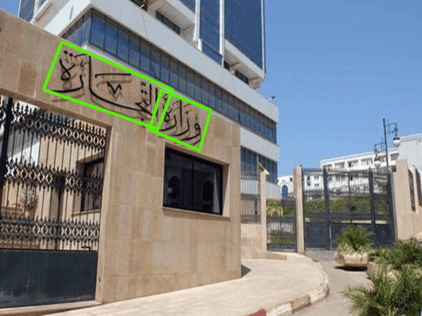}
\caption{}
\end{subfigure}
\caption{Illustration of the proposed scene text detection technique: Successful detections where sample images are picked from the four studied datasets including (a) MSRA-TD500, (b) ICDAR2013, (c) ICDAR2017-RCTW and (d) ICDAR2017-MLT, respectively.}
\label{fig:det_sample}
\end{figure}

The proposed technique is also evaluated on two more recent large-scale datasets including the ICDAR2017-RCTW and the ICDAR2017-MLT. As the two datasets both have a large amount of training images, we evaluated the proposed semantics-aware text borders only. As Table \ref{tab:icdar17_mlt} shows, the use of semantics-aware text borders helps to improve both detection recall and detection precision clearly for the ICDAR2017-RCTW dataset where a mixture of word level and text-line level annotations is created. For the ICDAR2017-MLT dataset, the improvement is mainly from higher precisions as most annotations in this dataset are at word level. In addition, the proposed techniques outperform state-of-the-art methods (including baselines and winning methods as reported in the ICDAR2017-RCTW and ICDAR2017-MLT benchmarking competition papers \cite{nayef2017icdar2017,shi2017icdar2017}) for both datasets when DenseNet is used.\medskip

\noindent\textbf{Qualitative Results.} Fig. \ref{fig:det_sample} shows several sample images and the corresponding detections by using the proposed technique, where all sample images are picked from the four studied datasets including several images in Figs. \ref{fig:aug_show} and \ref{fig:border_show}. As Fig. \ref{fig:det_sample} shows, the proposed technique is capable of detecting scene texts that have different characteristics and suffer from different types of degradation. In particular, the inclusion of the bootstrapping based augmentation helps to produce more complete detections though texts may not be localized accurately as illustrated in Fig. \ref{fig:aug_show}. On the other hand, the inclusion of the semantics-aware text borders helps to produce more accurate scene text localization though texts may be detected by multiple broken boxes as illustrated in Fig. \ref{fig:border_show}. The combination of bootstrapping based augmentation and semantics-aware text borders overcomes both constraints (broken detection and inaccurate localization) and produces more complete and accurate text detections as illustrated in Fig. \ref{fig:det_sample}.\medskip

\begin{figure}[t]
\centering
\begin{subfigure}[t]{0.325\textwidth}
\includegraphics[width=\textwidth]{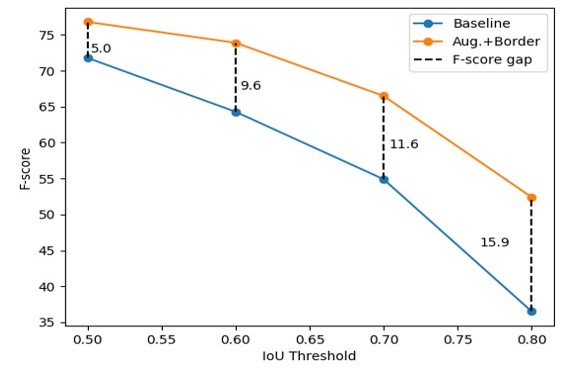}
\caption{}
\label{fig:iou}
\end{subfigure}
\begin{subfigure}[t]{0.325\textwidth}
\includegraphics[width=\textwidth]{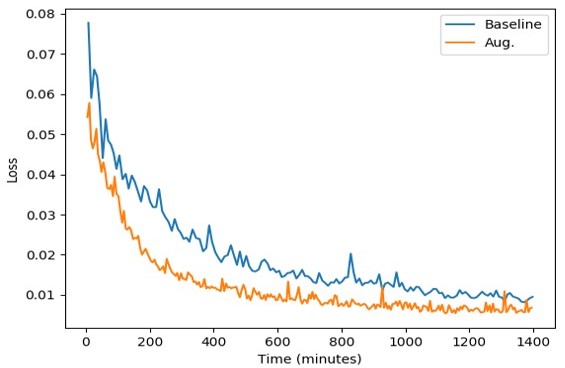}
\caption{}
\label{fig:time}
\end{subfigure}
\begin{subfigure}[t]{0.325\textwidth}
\includegraphics[width=\textwidth]{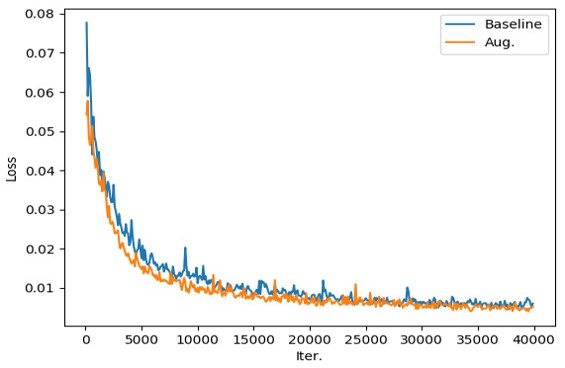}
\caption{}
\label{fig:iter}
\end{subfigure}
\caption{The inclusion of the semantics-aware text borders and bootstrapping based augmentation helps to improve the scene text localization accuracy greatly as illustrated in (a), where the f-score gap keeps increasing with the increment of IoU threshold on MSRA-TD500 dataset. Additionally, the inclusion of the bootstrapping based augmentation also leads to fast learning and convergence as illustrated in (b) and (c) on MSRA-TD500 dataset. }
\end{figure}

\noindent\textbf{Discussion.} The proposed technique is capable of producing accurate scene text localization which is critical to the relevant scene text recognition task. This can be observed in Fig. \ref{fig:iou} that shows f-scores of the proposed model (the semantics-aware text borders and augmented images are included in training) vs the baseline model (training uses the original images only) when different IoUs (Intersection over Union) are used in evaluation. As Fig. \ref{fig:iou} shows, the f-score gap increases from 5.0 to 15.9 steadily when the IoU threshold increases from 0.5 to 0.8, demonstrating more accurate scene text localization by the proposed technique. 

\begin{figure}[t]
\centering
\begin{subfigure}[t]{0.32\textwidth}
\includegraphics[width=\textwidth]{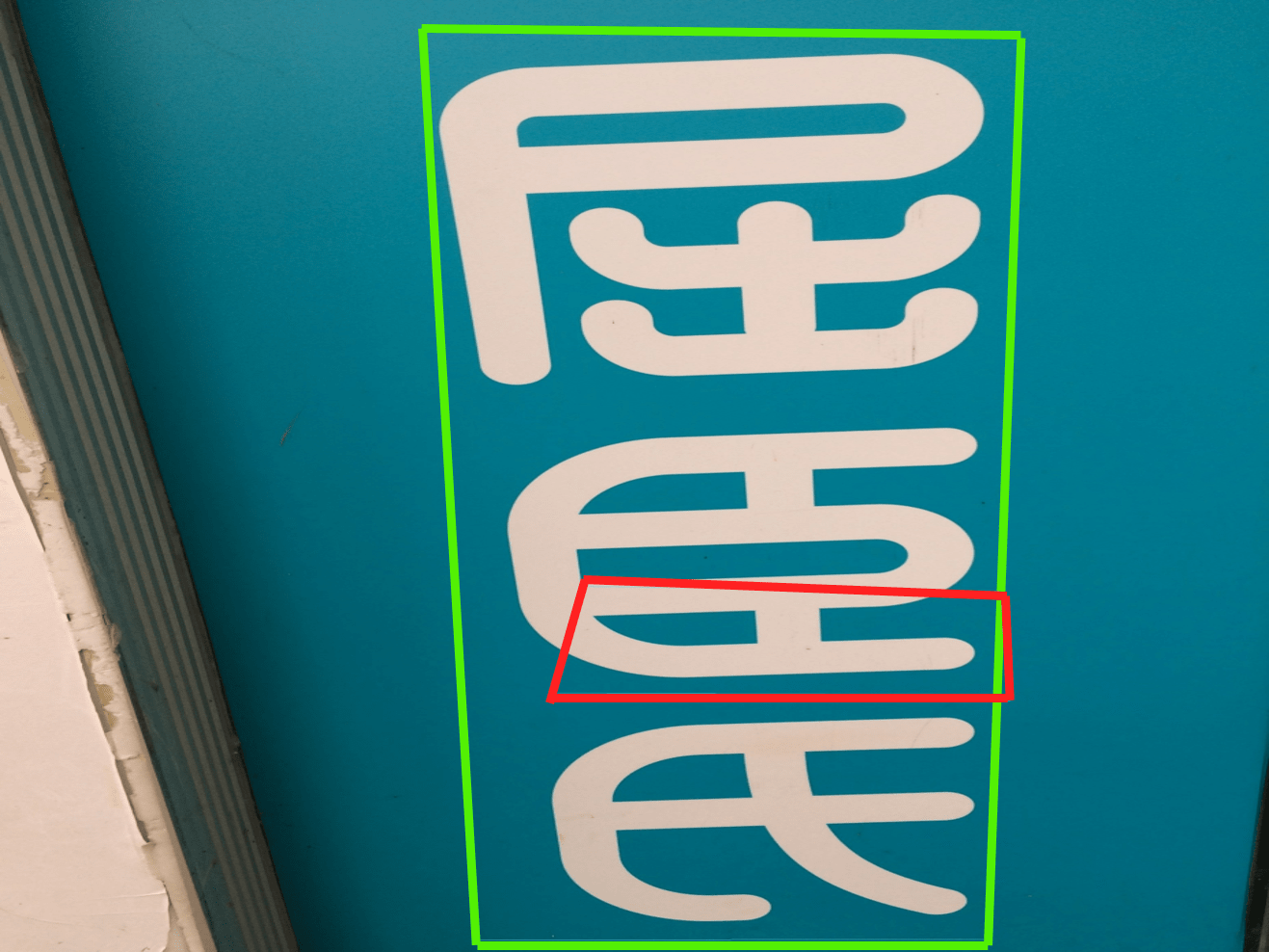}
\end{subfigure}
\begin{subfigure}[t]{0.32\textwidth}
\includegraphics[width=\textwidth]{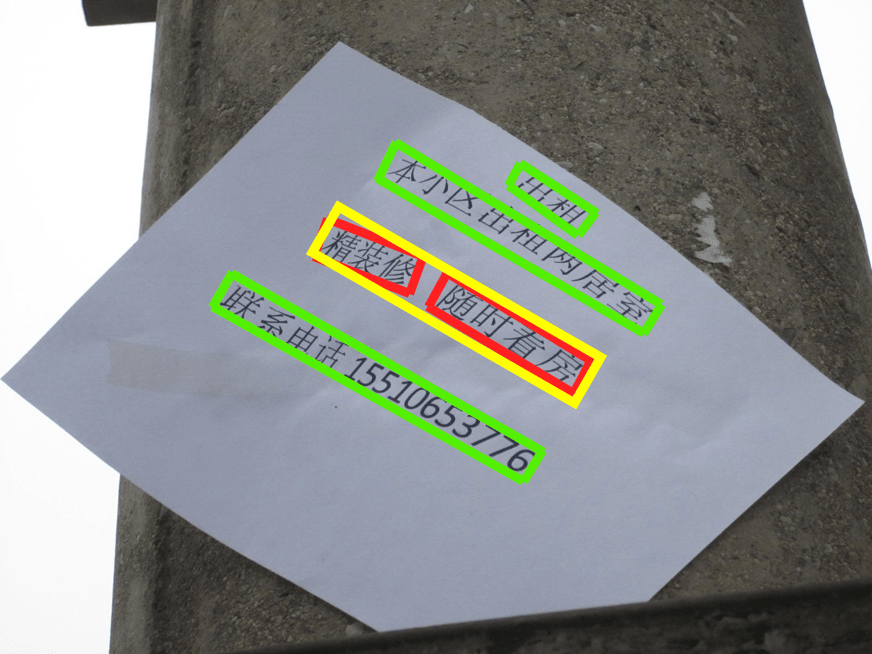}
\end{subfigure}
\begin{subfigure}[t]{0.32\textwidth}
\includegraphics[width=\textwidth]{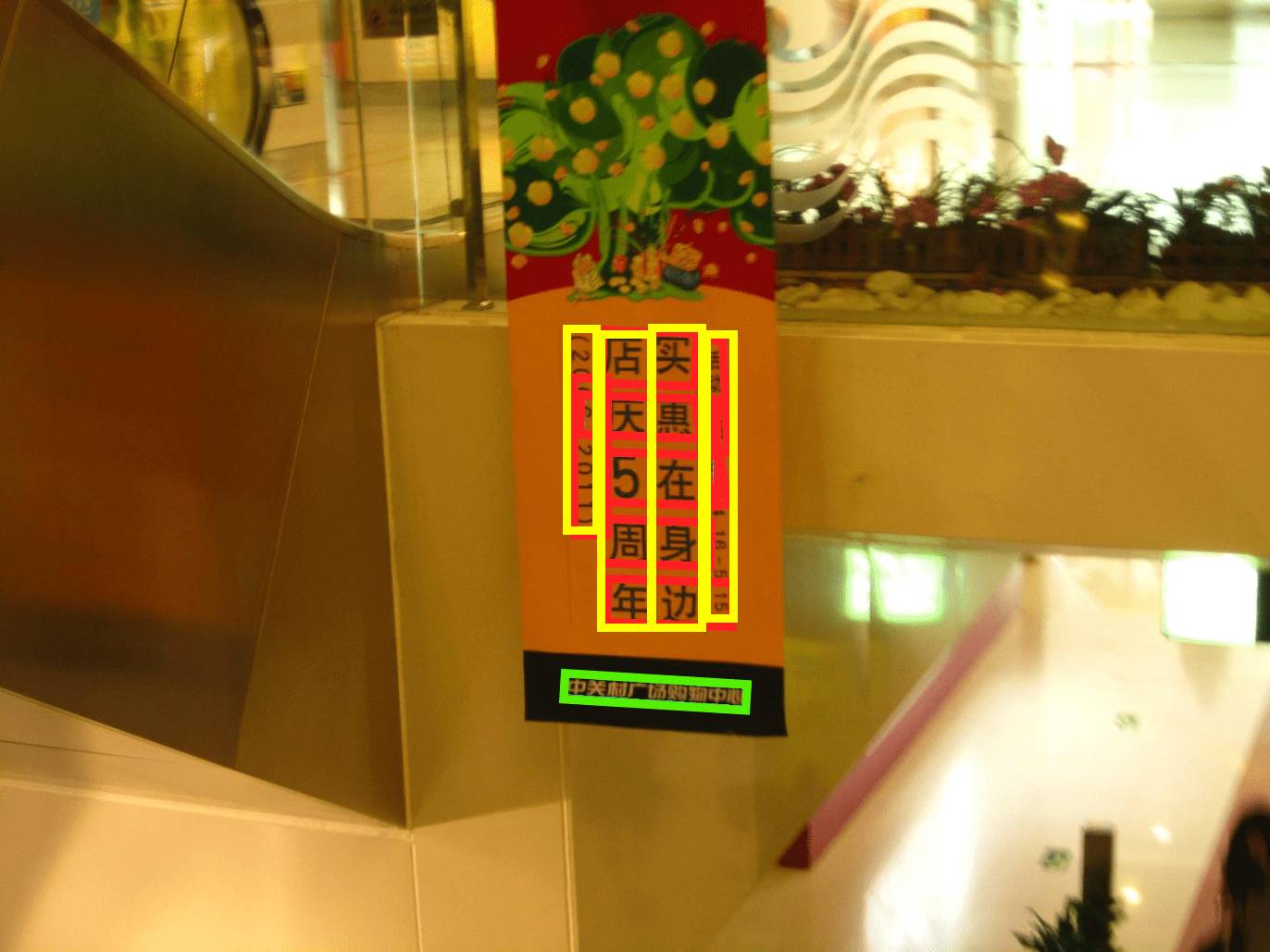}
\end{subfigure}
\caption{Illustration of the failure cases of proposed scene text detection technique: Sample images are from the four studied datasets, where green boxes are correct outputs of our methods, red boxes are false detections and yellow boxes give the ground-truth missing detections.}
\label{fig:fail}
\end{figure}

Another interesting observation is that the inclusion of the augmented images often accelerates the training convergence as illustrated in Fig. \ref{fig:time}. For training over 40,000 iterations (batch size of 16) on the MSRA-TD500 dataset, the model using the augmented images takes 36 hours while the baseline model using the original training images only takes 56 hours when both models converge and obtain f-scores of 74.3 and 71.8, respectively, as shown in Table \ref{tab:msra_icdar13}. This can be further verified by checking the training loss vs training iteration number as shown in Fig. \ref{fig:iter}. Experiments on other datasets show similar convergence pattern as shown in Figs. \ref{fig:time} and \ref{fig:iter}. This does not make sense at the first sight as the augmentation increases the number of training images by 20 times (20 augmented images are sampled for each training image). We conjecture that the faster convergence is largely due to the augmented text line segments that are shorter than the original text lines and accordingly decouple different types of image variation which leads to the faster learning and model convergence. 

The proposed technique could fail under several typical scenarios as illustrated in Fig. \ref{fig:fail}. First, it may introduce false positives while handling scene texts of a big size, largely due to NMS errors as shown in the first image. Second, it may produce incorrect broken detections when a text line has large blanks as shown in the second image. This kind of failure often results from annotation inconsistency where some long text line with large blanks is annotated by a single box whereas some is annotated by multiple boxes. Third, it could be confused when vertical texts can also be interpreted as horizontal and vice versa as shown in the third image. Without the text semantic information, it is hard to tell whether it is two vertical text lines or five horizontal words.

\section{Conclusions} \label{sec:conclusion}
This paper presents a novel scene text detection technique that makes use of semantics-aware text borders and bootstrapping based text segment augmentation. The use of semantics-aware text borders helps to detect text border segments with different semantics which improves the scene text localization accuracy greatly. The use of augmented text line segments helps to improve the consistency of predicted feature maps which leads to more complete instead of broken scene text detections. Experiments over four public datasets show the effectiveness of the proposed techniques.

\section{Acknowledgement}
This work is funded by the Ministry of Education, Singapore, under the project ``A semi-supervised learning approach for accurate and robust detection of texts in scenes" (RG128/17 (S)).

\clearpage

\bibliographystyle{splncs04}
\bibliography{egbib}
\end{document}